\documentclass[11pt]{article}

\pdfoutput=1

\usepackage[final]{acl}

\usepackage{booktabs}
\usepackage{array}
\usepackage{times}
\usepackage{latexsym}
\usepackage{amsmath,amsfonts,bm,bbm,mathtools,mleftright,nicefrac, amssymb}

\usepackage[T1]{fontenc}
\usepackage[utf8]{inputenc}
\usepackage{microtype}

\usepackage{inconsolata}

\usepackage{graphicx}
\usepackage{embeddings}
\usepackage{booktabs}

\usepackage[capitalize,noabbrev]{cleveref}
\crefname{section}{\S}{\S\S}
\crefname{table}{Table}{Tables}
\crefname{figure}{Figure}{Figures}
\crefname{equation}{Eq.}{Eq.}
\crefname{appendix}{App.}{}
\makeatletter
\AddToHook{cmd/appendix/before}{\def\cref@section@alias{appendix}}
\makeatother

\title{Probing for Reading Times}

\author{
  \textbf{Eleftheria Tsipidi\textsuperscript{\normalfont1}} \quad
  \textbf{Samuel Kiegeland\textsuperscript{\normalfont1}} \quad
  \textbf{Francesco Ignazio Re\textsuperscript{\normalfont1}} \quad
  \textbf{Tianyang Xu\textsuperscript{\normalfont2}} \\
  \textbf{Mario Giulianelli\textsuperscript{\normalfont3}} \quad
  \textbf{Karolina Sta\'nczak\textsuperscript{\normalfont1}} \quad
  \textbf{Ryan Cotterell\textsuperscript{\normalfont1}}
   \\
  \textsuperscript{1}ETH Zürich \quad
  \textsuperscript{2}Toyota Technological Institute at Chicago \quad
  \textsuperscript{3}University College London\\
  \texttt{\{\href{mailto:eleftheria.tsipidi@inf.ethz.ch}{eleftheria.tsipidi},
  \href{mailto:francesco.re@inf.ethz.ch}{francesco.re},
  \href{mailto:ryan.cotterell@inf.ethz.ch}{ryan.cotterell}\}@inf.ethz.ch}\\
  \texttt{\href{mailto:samuel.kiegeland@gmail.com}{samuel.kiegeland@gmail.com} \quad
  \href{mailto:sallyxu@ttic.edu}{sallyxu@ttic.edu}}\\
  \texttt{\href{mailto:m.giulianelli@ucl.ac.uk}{m.giulianelli@ucl.ac.uk}
\quad \href{mailto:karolinaewa.stanczak@ai.ethz.ch}{karolinaewa.stanczak@ai.ethz.ch}}
}

\begin{document}
\maketitle
\begin{abstract}
Probing has shown that language model representations encode rich linguistic information, but it remains unclear whether they also capture cognitive signals about human processing. In this work, we probe language model representations for human reading times. Using regularized linear regression on two eye-tracking corpora spanning five languages (English, Greek, Hebrew, Russian, and Turkish), we compare the representations from every model layer against scalar predictors---surprisal, information value, and logit-lens surprisal.
We find that the representations from early layers outperform surprisal in predicting early-pass measures such as first fixation and gaze duration. The concentration of predictive power in the early layers suggests that human-like processing signatures are captured by low-level structural or lexical representations, pointing to a functional alignment between model depth and the temporal stages of human reading.
In contrast, for late-pass measures such as total reading time, scalar surprisal remains superior, despite its being a much more compressed representation. We also observe performance gains when using both surprisal and early-layer representations. Overall, we find that the best-performing predictor varies strongly depending on the language and eye-tracking measure.\looseness-1

\vspace{.5em}
\hspace{1.25em}\includegraphics[width=1.25em,height=1.25em]{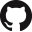}{\hspace{.75em}\parbox{\dimexpr\linewidth-2\fboxsep-2\fboxrule}{\url{https://github.com/rycolab/llm-representations-rt}}}
\end{abstract}

\section{Introduction}

How long a reader's eyes linger on a linguistic unit is posited to reflect the cognitive effort required to process it \citep{just1980theory, rayner1998eye}. One prominent way to measure these durations is eye-tracking, which records fixation times at fine temporal resolution.
A key question in psycholinguistics is which textual features best predict these reading times, and the predictive fit of a feature set serves as a measure of its \emph{psychometric power} \citep{SMITH2013302}. To date, the most successful neural language model-based predictor has been surprisal \citep{hale-2001-probabilistic, levy2008expectation, wilcox-2023-testing}.\looseness-1

Independently, a large body of work on \emph{probing} has demonstrated that the internal representations of neural language models encode a wealth of linguistic information, including syntactic structure, morphological features, and semantic properties \citep{alainbengio2017, white-etal-2021-non, immer-etal-2022-probing, kim2025linear}.
Yet probing studies have overwhelmingly focused on predicting properties of the linguistic signal itself from representations.
While recent work has shown that language model representations align with neural signals measured via fMRI and EEG \citep{schrimpf2021neural, caucheteux2022brains}, it remains unclear to what extent the language model's internal representations can directly predict \emph{behavioral} reading times---the fine-grained, unit-level processing effort that readers expend, as reflected in eye-tracking measures.\looseness-1

\begin{figure*}[ht]
    \centering
    \includegraphics[width=\linewidth]{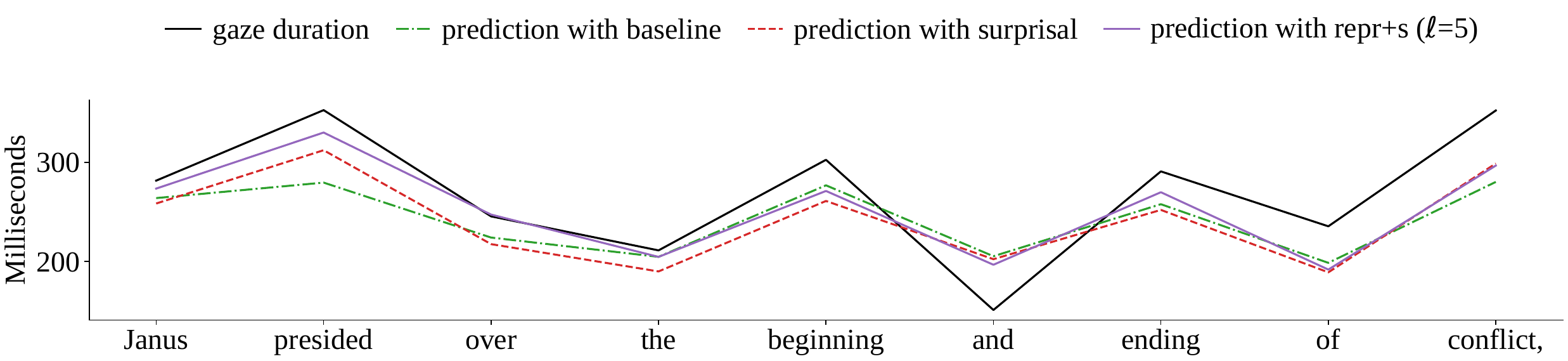}
    \caption{Gaze duration and its prediction by different mGPT-derived feature settings. The excerpt is from a document in the MECO dataset. The y-axis represents reading time measured in milliseconds. True gaze duration is represented by a black line. The purple line represents the predictions of a linear model trained on 5th-layer representations and standard surprisal. Note how the gaze duration and its predictions spike on units with high information content, such as \textit{presided} and \textit{conflict}.}\looseness=-1
    \label{fig:meco-example}
\end{figure*}

In this work, we probe language model representations for human reading times. Using regularized linear regression, we predict unit-level reading times directly from the neural language models' representations extracted at every layer of a language model. We compare these representation-based predictors against scalar baselines---surprisal, information value \citep{giulianelli-etal-2024-generalized}, and logit-lens surprisal \citep{nostalgebraist2020logitlens, kuribayashi2025large}---which compress the model's internal state into a single dimension.
An illustration of this predictive task is provided in \cref{fig:meco-example}, which shows the true aggregated unit-by-unit gaze duration of human readers and the gaze duration predicted by various predictor variables.
We conduct our evaluation on two eye-tracking corpora, Provo \citep{provo} and MECO \citep{siegelman2022expanding}, spanning five languages: English, Greek, Hebrew, Russian, and Turkish, using mGPT \citep{shliazhko-etal-2024-mgpt}, GPT-2 \citep{radford2019language}, and cosmosGPT \citep{kesgin2024introducing}. We evaluate the predictive power of representations from all layers for three reading time metrics: first fixation duration, gaze duration, and total reading time.\looseness=-1

Our results reveal clear differences across reading time modalities.
In English, representations from early layers tend to outperform surprisal in predicting early-pass measures, such as first fixation duration and gaze duration, suggesting that features relevant to initial lexical access and local structural encoding are accessible in internal states beyond what surprisal captures. In contrast, for late-pass measures such as total reading time, scalar predictors, especially surprisal and logit-lens surprisal, are often competitive with or superior to high-dimensional representations. We also observe substantial cross-lingual variation in the relative predictive power of scalar and representation-based predictors. In Greek, Hebrew, Russian, and Turkish, scalar predictors are frequently as strong as or stronger than representations, depending on the eye-tracking measure. We further find that combining surprisal with layer-wise representations frequently improves predictive performance over representations alone, although the gains over scalar baselines are less consistent. Overall, our findings show that the psychometric power of language models depends strongly on the reading-time measure, the model layer, and the language under study, rather than being captured by a single predictor across all settings. \looseness-1

\section{Preliminaries}

\paragraph{Language Models.}
We adopt the formulation of \citet{kiegeland-etal-2026-units}, who distinguish the abstract linguistic \defn{units} that humans process, over which reading times are modeled, and \defn{symbols}, which the language model outputs. Throughout this section, we present surprisal theory and our predictors in terms of units. We discuss how to reconcile this formulation with a language model defined over tokens in \cref{sec:feature-estimation}.
Let $\alphabet$ be a countable set of units.
A \defn{string} $\str = \sym_1 \dots \sym_T$ is a finite sequence of units $\sym_t \in \alphabet$. We write $\str_{<t} = \sym_1 \cdots \sym_{t-1}$ for the prefix of $\str$ up to but not including position~$t$. 
We denote string concatenation by juxtaposition, i.e., $\str \str'$ denotes the concatenation of $\str$ and $\str'$.
With $\Strings$, we denote the Kleene closure of $\alphabet$, i.e., the set of all finite strings over $\alphabet$. 
A \defn{language model} is a probability distribution $\plm$ over $\Strings$.
Every language model $\plm$ induces a \defn{prefix probability}, defined as
\begin{equation}
\prefixlm(\str) \defeq \sum_{\str' \in \Strings} \plm(\str \str').
\end{equation}
Then, define the \defn{conditional prefix probability} as\looseness=-1
\begin{equation}
\prefixlm(\sym \mid \str) \defeq \frac{\prefixlm(\str \sym)}{\prefixlm(\str)}.
\end{equation}
By the chain rule of probability, the language model factorizes autoregressively as
\begin{equation}
\plm(\str) = \prefixlm(\eos \mid \str) \prod_{t=1}^{T} \prefixlm(\sym_t \mid \str_{<t}),
\end{equation}
where $\eos$ is a distinguished end-of-string symbol. Let $\eosalphabet \defeq  \alphabet\cup \{\eos\}$.\looseness-1

\paragraph{Neural Language Models.}
Modern language models, such as those based on the transformer architecture \citep{vaswani2017attention}, parameterize the conditional distributions above through a stack of $L$ layers.
The input layer maps each symbol $\sym \in \eosalphabet$ to a vector $\emb_{0}(\sym) \in \R^{\embdim}$, and each subsequent layer computes a representation as a function of the previous layer's representations.
Let $\str = \sym_1 \cdots \sym_T$ be a string over $\eosalphabet$, then for each layer $\layer \in \{1, \ldots, L\}$, we define
\begin{equation}
\emb_{\layer}(\str) \defeq \layerfn_{\layer}\left(\emb_{\layer-1}(\sym_{1}), \dots, \emb_{\layer-1}(\sym_{T})\right),
\end{equation}
where $\emb_{\layer}(\str)$ denotes the $\layer$-layer representation at the final unit position $T$ and $\layerfn_{\layer} \colon (\R^{\embdim})^{*} \to \R^{\embdim}$ denotes the transformation at layer $\layer$. 
The last layer representation is then projected onto $\Delta(\eosalphabet)$ as\looseness=-1
\begin{equation}
\label{eq:lm-output}
\prefixlm(\cdot \mid \str) = \softmax\left(\unemb \outputfn(\emb_L (\str)) + \ubias\right),
\end{equation}
where $\outputfn\colon\R^{\embdim} \to \R^{\embdim}$ is a final (non-linear) transformation applied before the linear projection (e.g., layer normalization), $\unemb \in \R^{(|\eosalphabet|) \times \embdim}$ is the projection matrix, and $\ubias \in \R^{|\eosalphabet|}$ is a bias term.
The final representation $\emb_L$ is directly used to compute the distribution over the next symbols.\footnote{
    Parameterized LMs typically include a special end-of-sequence token, for which representations can be computed and on which next-token predictions can in principle be conditioned. From the perspective of an LM as a distribution over strings, however, conditioning on $\eos$ is not well defined. In this work, $\eos$ may only appear as the final unit of a string $\str$, where it is used to model wrap-up effects (see \cref{sec:psychometric-data}).
}
However, the intermediate representations $\emb_1(\str), \ldots, \emb_{L-1}(\str)$ encode linguistic information themselves \citep{alainbengio2017, immer-etal-2022-probing}. In this work, we investigate whether these representations also encode information predictive of human reading behavior.

\section{Psychometric Data} \label{sec:psychometric-data}
In this work, we study how well we can predict real-valued measurements of human processing effort collected during natural reading. Formally, for a unit $\sym_t \in \eosalphabet$ read in context $\str_{<t} \in \Strings$, we observe a reading time $\readingtime(\sym_t, \str_{<t}) \in \R$; when $\sym_t = \eos$, this corresponds to utterance-final \defn{wrap-up} cost \citep{rayner2000wrapup, meister-etal-2022-analyzing}, reflecting the additional processing cost associated with integrating the full utterance.
Eye-tracking experiments yield several such measurements per unit, corresponding to different stages of processing: first-fixation duration (the duration of the initial fixation), gaze duration (the sum of all fixations before the eyes leave the unit), and total reading time (the sum of all fixations including regressions). Our goal is to predict these reading times from features derived from a language model.

\subsection{Previously Proposed Predictors}
We now discuss three previously proposed \emph{scalar} predictors of human reading time.
\paragraph{Surprisal Theory.}
Surprisal theory \citep{hale-2001-probabilistic, levy2008expectation} posits that reading times are an affine function of \defn{surprisal}, the negative log-probability of a unit under the reader's implicit language model. Formally, let $\pHum$ denote the \defn{human language model}---the probability distribution that characterizes a reader's expectations over upcoming linguistic material. 
The \defn{surprisal} of unit $\sym_t$ in context $\str_{<t}$ is then
\begin{equation}
\surprisal(\sym_{t}, \str_{<t}) \defeq - \log \pHum (\sym_t \mid \str_{<t}),
\end{equation}
and the theory predicts that reading time is an affine function of this quantity \citep{SMITH2013302, shain2024evidence}. Since $\pHum$ is not directly observable, it is standard practice to approximate it with a trained language model $\plm$, and empirical support for the resulting predictions has been found across diverse datasets and languages \citep{wilcox-2023-testing}. Note that the predictive power of $\plm$-derived surprisal depends on how well $\plm$ approximates $\pHum$, and this approximation quality likely varies across languages and models.\looseness-1

\paragraph{Information Value.}
Shannon surprisal is the standard metric for quantifying the unexpectedness of a linguistic unit under a model $\plm$, but other operationalizations of information content exist; for an overview, see \citet{giulianelli-etal-2024-generalized}.
In this paper, we include next-unit information value in addition to standard surprisal.
Next-unit information value measures the expected distance between the observed next unit $\sym_t$ and alternative continuations $\sym \in \eosalphabet$ sampled from the model's predictive distribution. 
This corresponds to a special case of the general string-level formulation of information value, where continuations are restricted to a single unit \citep[cf.][]{giulianelli-etal-2023-information,giulianelli-etal-2024-generalized}.
Formally, it is defined as 
\begin{align}
    \infoval(\sym_t, \str_{<t}) \defeq \Expect_{\sym \sim \prefixlm(\cdot \mid \str_{<t})} \left[ \distance(\sym_t, \sym) \right], \label{eq:information-value}
\end{align}
where $\distance \colon \eosalphabet \times \eosalphabet \to \R_{\geq 0}$ is a distance function, typically operationalized as the cosine distance between the contextual representations $\emb_{\layer}(\str_{<t}\sym_t)$ and $\emb_{\layer}(\str_{<t}\sym)$ at a given layer $\layer$ 
\citep{giulianelli-etal-2024-generalized,giulianelli-etal-2024-incremental}. This makes information value a natural point of comparison for our representation-based predictors.

\paragraph{Logit Lens.}

Standard surprisal is computed from the final layer's next-token distribution. The \defn{logit lens} \citep{nostalgebraist2020logitlens, kuribayashi2025large} asks what distribution an intermediate layer would induce if its representation were fed directly to the output head. Concretely, it applies the \emph{same} projection matrix $\unemb$, bias $\ubias$, and layer normalization $\layernorm$ that are used after the final layer to the representation of an earlier layer $\layer$:
\begin{equation}
\distLogitlens_{\layer}(\cdot \mid \str)
\defeq
\softmax\left(\unemb\emb_{\layer}(\str) + \ubias\right).\footnote{Even at the final layer $\layer = L$, the logit-lens distribution $\distLogitlens_L$ need not equal the model's true output distribution $\prefixlm$, because \cref{eq:lm-output} includes the non-linear transformation $\outputfn$ (e.g., layer normalization) which is absent from the logit lens.}
\label{eq:logitlens_dist}
\end{equation}
Because $\unemb$ and $\ubias$ are estimated only to decode the final layer's representation, there is no guarantee that this projection yields a meaningful distribution at earlier layers---the intermediate representations may not be linearly decodable in vocabulary space. In practice, however, the logit lens has been found to produce interpretable predictions at many layers \citep{nostalgebraist2020logitlens}.
We define the \defn{logit-lens surprisal} $\logitlens$ at layer $\layer$ as
\begin{equation}
\logitlens[\layer](\sym_t,\str_{<t})
\defeq
-\log \distLogitlens_{\layer}(\sym_t \mid \str_{<t}).
\label{eq:logitlens_surp}
\end{equation}

\subsection{The Limitations of Scalar Predictors}

All three predictors introduced above---surprisal, information value, and logit-lens surprisal---share a fundamental limitation: each compresses a representation extracted from a language model into a single scalar. 
While such scalar predictors have served as useful proxies for human processing effort, it is natural to suspect that using the entire representation as a predictor may be more useful.
Moreover, in the case of surprisal, larger models that achieve lower cross-entropy on held-out text often evince poorer fit to human reading times \citep{oh-schuler-2023-why, kuribayashi-etal-2024-psychometric}, and recent fine-grained modeling suggests that much of the variance typically attributed to surprisal may instead be explained by skip rates \citep{re-etal-2025-spatio}.
Finally, recent evidence also indicates that the internal layers of language models contain representations that align more closely with human behavioral and neural signals than any single scalar derived from them \citep{schrimpf2021neural, caucheteux2022brains, kuribayashi2025large}.
Taken together, this suggests that scalar compression---whether through surprisal (which reduces the final layer to a log-probability), information value (which summarizes representational distance as a single expectation), or logit-lens surprisal (which projects an intermediate layer through the output head)---discards much of the psychometrically relevant information contained in the model's internal representations.

\section{Methods}
To evaluate whether the representations induced by neural language models serve as useful predictors of human processing effort, we apply various forms of regularized linear regression.
By controlling for standard psycholinguistic factors, we compare the predictive power of representations, information value, standard surprisal, and layer-wise surprisal.

\subsection{Linear Regression}
\label{sec:method-model}

To predict reading times, we follow standard psycholinguistic practices and use linear models \citep{goodkind2018predictive, wilcox-etal:2020-on-the-predictive-power}.
Formally, let $\readingtime(\sym_t, \str_{<t}) \in \R$ be a real-valued reading time measurement for a unit $\sym_t \in \eosalphabet$ in context $\str_{<t} \in \Strings$, and let $\predvec(\sym_t, \str_{<t}) \in \R^D$ be a column vector of predictor variables.\footnote{All vectors in this paper are column vectors.} We predict reading times as
\begin{equation}
\label{eqn:predreadingtime}
\predr(\sym_t, \str_{<t}) \defeq \predvec(\sym_t, \str_{<t})^\top \betaRegTrg,
\end{equation}
where $\betaRegTrg \in \R^D$ is a parameter vector. Let the corpus consist of $N$ strings $\str^{(1)}, \dots, \str^{(N)}$, where string $\str^{(n)}$ has $T^{(n)}$ units plus $\eos$. We estimate $\betaRegTrg$ by minimizing the per-string squared loss:
\begin{multline}
\label{eqn:per_string_loss}
L_n(\betaRegTrg)
\defeq
\sum_{t=1}^{T^{(n)}}
\bigl(\readingtime(\sym_t^{(n)}, \str_{<t}^{(n)}) - \predr(\sym_t^{(n)}, \str_{<t}^{(n)})\bigr)^2 \\
+ \bigl(\readingtime(\eos, \str^{(n)}) - \predr(\eos, \str^{(n)})\bigr)^2.
\end{multline}
Ordinary least squares estimates $\betaRegTrg$ by minimizing $L(\betaRegTrg) \defeq \sum_{n=1}^{N} L_n(\betaRegTrg)$.
Following \citet{wilcox-2023-testing} and \citet{opedal-etal-2024-role}, we do not apply any transformation (e.g., log or $z$-score) to the reading times before fitting the model, so that $\predr$ is directly interpretable in milliseconds.\looseness=-1

\paragraph{Regularized Linear Regression.}
We also consider regularized variants. 
Ridge regression adds an $\lVert \cdot \rVert_2^2$ penalty:
\begin{equation}
L_{\mathrm{R}}(\betaRegTrg)
\defeq L(\betaRegTrg)
+
\penaltyWeight \lVert \betaRegTrg \rVert_2^2,
\end{equation}
where $\penaltyWeight \ge 0$ controls the strength of regularization.
LASSO regression instead uses an $\lVert \cdot \rVert_1$ penalty:
\begin{equation}
L_{\mathrm{L}}(\betaRegTrg)
\defeq L(\betaRegTrg)
+
\penaltyWeight \lVert \betaRegTrg \rVert_1.
\end{equation}
In contrast to ridge regression, LASSO encourages sparsity in $\betaRegTrg$, inducing sparse solutions and acting as a form of feature selection.\looseness=-1

\paragraph{Tuning.}
\label{ssec:param-tuning}
We tune the regression models by selecting (i) whether to apply regularization and, if so, (ii) whether to use LASSO or ridge regression, along with (iii) the corresponding penalty weight. Model selection is performed using the test \defn{mean squared error} (MSE) on a fixed train--test split:
\begin{equation}
\label{eqn:mse}
\mathrm{MSE}(\betaRegTrg) \defeq \frac{L(\betaRegTrg)}{\sum_{n=1}^{N}(T^{(n)}+1)}.
\end{equation}
To avoid leakage, the documents used in this tuning test split (5 documents in Provo and 2 documents in MECO) are excluded from all subsequent experiments.
We evaluate penalty weights in the range $[0.001, 10]$, performing hyperparameter selection independently for each predictor type, layer, and dependent variable. This procedure is applied to the baseline and surprisal models, and to each layer-wise instance (layers 1--24 for mGPT and 1--12 for GPT-2 and cosmosGPT) of information value, logit lens, and representation predictors.

\paragraph{Cross-Validation.} 
We evaluate each combination of predictor type and reading time measure using 10-fold cross-validation, run separately on Provo and on each language subset of MECO.\looseness-1

\section{Experimental Setup}

\subsection{Feature Estimation}
\label{sec:feature-estimation}
\paragraph{Models.}
We use surprisal estimates from mGPT \citep{shliazhko-etal-2024-mgpt}, a multilingual model based on the GPT-3 \citep{brown-etal-2020-gpt3} architecture. mGPT was trained on 61 languages from 25 language families, which enables us to experiment on both Provo \citep{provo} and the MECO \citep{siegelman2022expanding} data. It has 24 layers, each with an embedding dimension of 2048.
For additional experiments, we use two monolingual models: the English monolingual GPT-2 Small \citep{radford2019language} on the Provo data and the English subset of MECO; and the Turkish cosmosGPT \citep{kesgin2024introducing} on the Turkish subset of MECO. Both monolingual models have 12 layers with an embedding dimension of 768.

\paragraph{From Tokens to Units.}
Let $\ptok$ denote the token-level\footnote{See \citet{gastaldi2025the} and \citet{pmlr-v267-vieira25a} for a formal treatment of tokenized and token-level language models.\looseness=-1} language model: a probability distribution over token strings $\tokAlphabet^{\ast}$, where $\tokAlphabet$ is a finite token alphabet, and $\tokenizer \colon \Strings \to \tokAlphabet^{\ast}$ is a function that maps a unit string to a token string. We assume that $\tokenizer$ respects unit boundaries: no token spans two units, so any tokenization decomposes as $\tokenizer(\sym_1\cdots\sym_T) = \tokstr_1\cdots\tokstr_T$, where $\tokstr_t = \tok_{t,1}\cdots\tok_{t,n_t}$ is the token sequence corresponding to unit $\sym_t$.\footnote{For the whitespace-delimited words used in our corpora and the token alphabets of mGPT, GPT-2, and cosmosGPT, this holds in practice: each token sits within a single word.} Following standard practice \citep{wilcox-2023-testing}, we define unit-level surprisal and logit-lens surprisal as the sum over $\tokstr_t$ of the per-token surprisals under $\ptok$ and $\distLogitlens_\layer$ (\cref{eq:logitlens_dist}), respectively.\footnote{
    See \citet{giulianelli-etal-2024-proper}, \citet{pimentel-meister-2024-compute}, \citet{oh-schuler-2024-leading}, and \citet{kiegeland-etal-2026-units} for discussion of this choice and alternative approaches to calculating unit-level surprisal.
} Unlike \citet{kuribayashi2025large}, we do not include tuned lens \citep{belrose2025tuned}, since mGPT does not have a pre-trained tuned-lens model; we include logit-lens surprisal of the last layer, as it may differ from standard surprisal.\footnote{The \href{https://huggingface.co/docs/transformers/en/main_classes/output}{Hugging Face documentation} states that some models do in fact apply a function $\outputfn$ or further processing to the last state when it is returned; this could affect surprisal and render it different from the corresponding final layer logit lens.}

Writing $\emb_\layer(\tokstr)$ for the layer-$\layer$ hidden state of $\ptok$ at the final token of a token string $\tokstr \in \tokAlphabet^{\ast}$, we define the unit-level representation of $\str_{<t}\sym_t$ as the mean over the $n_t$ tokens that correspond to $\sym_t$:
\begin{equation}
\emb_\layer(\str_{<t}\sym_t)\defeq n_t^{-1}\!\!\!\!\sum_{k=M-n_t+1}^{M}\!\!\!\!\emb_\layer(\tokenizer(\str_{<t}\sym_t)_{\leq k})
\end{equation}
Mean-pooling is one of several possible aggregations; we discuss this in the limitations section.
For information value, \Cref{eq:information-value} then applies with $\distance$ computed as the cosine distance between these pooled representations, and we approximate the expectation by Monte Carlo with $\budget = 50$ continuations sampled from $\ptok$.\footnote{For each continuation, we generate tokens until an end-of-unit marker is encountered, or 3 tokens have been produced.}

\begin{figure*}[t!]
    \centering
    \includegraphics[width=\textwidth]{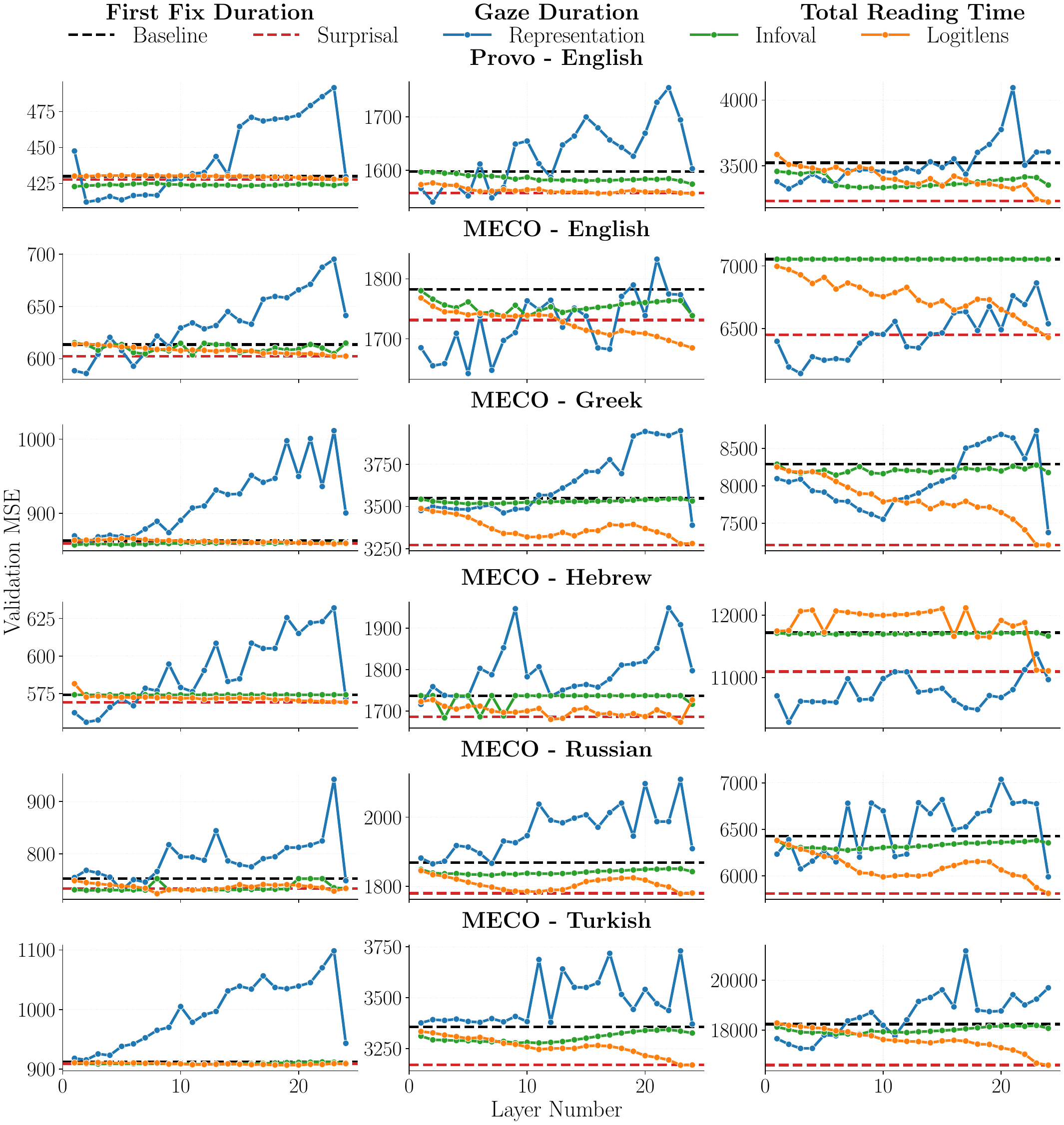}
    \caption{MSE for baseline, surprisal, representations, information value, and logit-lens surprisal on the Provo and MECO data across the 24 layers of mGPT and eye-tracking measures.
   \vspace{-15pt} 
    }
    \label{fig:mse_mgpt_layers}
\end{figure*}

\subsection{Data}
We use reading time data from
two commonly used corpora in psycholinguistics.
The Provo corpus \citep{provo} is a dataset of eye-movement behavior comprising eye-tracking recordings from 84 native English readers as they read 55 short English passages drawn from a range of fiction and nonfiction sources.
The Multilingual Eye Movement Corpus \citep[MECO;][]{siegelman2022expanding} is a large multilingual corpus with eye movement data from L1 speakers reading 12 Wikipedia-style passages in 13 languages. To capture a variety of different language families, we select the English, Greek, Hebrew, Russian, and Turkish portions of the dataset for our experiments.

\paragraph{Data Preprocessing.}
We quantify the unit-by-unit reading time, using three standard eye-tracking measures: \defn{first fixation duration}, the duration of the first fixation on a unit during first pass reading; \defn{gaze duration}, the sum of all consecutive fixations on the unit from first entry, until a fixation leaves it for the first time; and \defn{total reading time}, the sum of all fixations on the unit across the entire trial, including any later re-reading due to regressions.
In line with established psycholinguistic theory, data collected during eye-tracking experiments can be divided into early-pass and late-pass measures \citep{rayner1996mindless}. First fixation and gaze duration are considered early-pass measures, as they are primarily sensitive to the initial stages of unit recognition and lexical access \citep{rayner1998eye,cook2019eye}. Specifically, first fixation duration is viewed as a marker of initial orthographic and phonetic activation \citep{Rayner01082009}. Conversely, total reading time is a late-pass measure, which is interpreted as a marker of higher-level post-lexical processing, reflecting the cognitive effort required for syntactic integration, discourse-level comprehension, and the resolution of processing difficulties \citep{CLIFTON2007341}.\looseness=-1

\section{Results}
We now present results with surprisal, representations, information value, and logit lens computed using mGPT. For results using GPT-2 Small and cosmosGPT, see \Cref{app:monolingual}.
\subsection{Predictive Power of Representations}
\label{ssec:pred-power}
\Cref{fig:mse_mgpt_layers} and \Cref{tab:mse_mgpt} compare predictive power across layers and eye-tracking measures for Provo. Performance is highest for early-layer representations (1--10), declines in intermediate layers, and recovers at the final layer. The best representation layer is comparable to or stronger than surprisal, though surprisal sometimes wins despite its much lower dimensionality. Combining representations with surprisal improves over either alone; these gains tend to be significant over representations but not over scalars (\Cref{app:combined}).\looseness=-1

\begin{table*}[t!]
\centering
\small
\begin{tabular}{lrrrrr}
\toprule
Measure & Surprisal & Best $\emb\;(\layer)$ & Best $\infoval\;(\layer)$ & Best $\logitlens\;(\layer)$ \\
\midrule
\multicolumn{5}{c}{\textbf{Provo---English}} \\
\midrule
FFD & -2.28$_{4.55}$$^{*}$ & \textbf{-17.92$_{8.76}$$^{*\bullet}$ (2)} & -7.13$_{7.02}$$^{*}$ (1) & -2.31$_{4.54}$$^{*}$ (24) \\
GD & -39.96$_{34.34}$$^{*}$ & \textbf{-57.21$_{31.77}$$^{*}$ (2)} & -23.62$_{19.43}$$^{*}$ (24) & -41.34$_{35.16}$$^{*}$ (24) \\
TRT & -290.61$_{134.19}$$^{*\bullet}$ & -198.09$_{196.67}$$^{*\bullet}$ (2) & -189.52$_{100.75}$$^{*\bullet}$ (10) & \textbf{-298.87$_{135.03}$$^{*\bullet}$ (24)} \\
\midrule
\multicolumn{5}{c}{\textbf{MECO---English}} \\
\midrule
FFD & -11.09$_{18.93}$$^{*}$ & \textbf{-27.74$_{29.25}$$^{*}$ (2)} & -9.91$_{20.48}$$^{*}$ (11) & -11.29$_{18.60}$$^{*}$ (23) \\
GD & -50.98$_{120.48}$$^{*}$ & \textbf{-139.92$_{169.75}$$^{*\bullet}$ (5)} & -44.78$_{33.77}$$^{*\bullet}$ (10) & -97.28$_{112.91}$$^{*}$ (24) \\
TRT & -605.14$_{433.80}$$^{*\bullet}$ & \textbf{-914.85$_{746.53}$$^{*\bullet}$ (3)} & 0.00$_{0.00}$$^{*}$ (1) & -626.49$_{441.29}$$^{*\bullet}$ (24) \\
\midrule
\multicolumn{5}{c}{\textbf{MECO---Greek}} \\
\midrule
FFD & -3.75$_{9.13}$$^{*}$ & -4.72$_{22.70}$$^{*}$ (2) & \textbf{-6.10$_{10.71}$$^{*}$ (1)} & -4.67$_{10.32}$$^{*}$ (23) \\
GD & \textbf{-275.80$_{212.98}$$^{*\bullet}$} & -158.94$_{306.51}$$^{*}$ (24) & -32.10$_{38.16}$$^{*}$ (5) & -268.69$_{192.38}$$^{*\bullet}$ (23) \\
TRT & -1079.68$_{944.67}$$^{*\bullet}$ & -914.45$_{1072.60}$$^{*\bullet}$ (24) & -145.36$_{152.23}$$^{*\bullet}$ (6) & \textbf{-1080.86$_{939.97}$$^{*\bullet}$ (24)} \\
\midrule
\multicolumn{5}{c}{\textbf{MECO---Hebrew}} \\
\midrule
FFD & -5.04$_{5.08}$$^{*}$ & \textbf{-18.31$_{24.28}$$^{*}$ (2)} & 0.00$_{0.00}$$^{*}$ (1) & -4.84$_{4.68}$$^{*}$ (24) \\
GD & -50.97$_{58.09}$$^{*}$ & -20.96$_{114.62}$$^{*}$ (1) & -53.54$_{55.93}$$^{*}$ (3) & \textbf{-64.09$_{59.38}$$^{*\bullet}$ (23)} \\
TRT & -621.84$_{786.76}$$^{*\bullet}$ & \textbf{-1431.52$_{2384.96}$$^{*\bullet}$ (2)} & -53.61$_{32.12}$$^{*\bullet}$ (24) & -610.27$_{781.71}$$^{*\bullet}$ (24) \\
\midrule
\multicolumn{5}{c}{\textbf{MECO---Russian}} \\
\midrule
FFD & -18.74$_{22.90}$$^{\bullet}$ & -21.86$_{28.87}$$^{*}$ (5) & -22.31$_{26.95}$$^{*}$ (2) & \textbf{-28.75$_{27.04}$$^{*\bullet}$ (8)} \\
GD & -88.70$_{110.92}$$^{*\bullet}$ & -3.59$_{101.50}$$^{*}$ (2) & -36.36$_{38.45}$$^{*}$ (7) & \textbf{-89.84$_{102.91}$$^{*\bullet}$ (23)} \\
TRT & \textbf{-619.68$_{741.79}$$^{*\bullet}$} & -438.41$_{605.05}$$^{*\bullet}$ (24) & -153.04$_{149.30}$$^{*}$ (7) & -615.77$_{735.20}$$^{*\bullet}$ (24) \\
\midrule
\multicolumn{5}{c}{\textbf{MECO---Turkish}} \\
\midrule
FFD & -2.63$_{6.62}$ & 3.67$_{29.35}$$^{*}$ (2) & -3.79$_{9.74}$ (12) & \textbf{-5.21$_{6.39}$$^{*}$ (19)} \\
GD & -186.91$_{119.24}$$^{*\bullet}$ & 14.06$_{167.93}$$^{*}$ (24) & -78.60$_{47.64}$$^{*\bullet}$ (11) & \textbf{-187.66$_{118.60}$$^{*\bullet}$ (23)} \\
TRT & -1642.88$_{1410.20}$$^{*\bullet}$ & -973.47$_{1871.74}$$^{*\bullet}$ (3) & -403.02$_{217.54}$$^{*\bullet}$ (6) & \textbf{-1644.42$_{1409.93}$$^{*\bullet}$ (24)} \\
\bottomrule
\end{tabular}
    \caption{$\dmse$ (baseline--target) of ten-fold cross-validation for models trained on baseline features and mGPT-derived surprisal, representations ($\emb$), information value ($\infoval$), and logit-lens surprisal ($\logitlens$) on the Provo and MECO data across the 24 layers of mGPT and eye-tracking measures. For each measure, we report the lowest MSE over layers and the corresponding layer index $\layer$. Bold indicates the best condition per row. Asterisks (*) denote models that significantly outperform the respective models trained on permuted reading times, according to a one-sided paired $t$-test ($\significance = 0.001$). Similarly, bullets ($\bullet$) indicate significance over the baseline.}
    \label{tab:mse_mgpt}
\end{table*}

\subsection{Information Value and Logit Lens}
\label{ssec:information-value}
\Cref{fig:mse_mgpt_layers} and \Cref{tab:mse_mgpt} compare the predictive power of information value and logit-lens surprisal against model representations and surprisal. Overall, these predictors show less variability across layers compared to model representations. For first fixation duration and total reading time, we find that information value is more predictive when computed from the early to intermediate layers than from the later ones. In contrast, logit-lens surprisal tends to perform best in the last layers.\looseness=-1

\subsection{Predictive Power across Languages}
To assess the predictive power of representations across languages, we repeat the experiments for five languages from the MECO dataset: English, Greek, Hebrew, Russian, and Turkish. 
Consistent with the results for Provo, \Cref{fig:mse_mgpt_layers,tab:mse_mgpt} show that representations from early layers are the most predictive, with performance decreasing in intermediate layers before recovering in the final layer. An exception to this is gaze duration and total reading time for Greek, as well as total reading time for Turkish, where the representations from the last layer perform best. Similarly, we find that information value is most predictive at early and intermediate layers, while logit-lens surprisal has the highest predictive power at later layers. Moreover, we observe that Russian and Turkish are the only languages where the combination of representation and surprisal (\Cref{tab:mse_mgpt_combo}) is not the best predictor for first-fixation duration and gaze duration. Overall, we find that the most predictive predictor varies strongly depending on the language and eye-tracking measure.\looseness-1

\subsection{Permuting Reading Times}
To further test whether the observed predictive performance reflects a meaningful relationship between the predictors derived from language models and the reading time data, we conduct a permutation test on the dependent variable. Specifically, we refit the same models from \cref{ssec:pred-power}, on training sets with their reading times randomly permuted across words, but without permuting the reading times of the respective validation sets. There are similarities to non-permuted trends across layers for representations; 
however, mean squared error scores are higher  across the board, and differences between the various sets of predictors tend to be smaller (\Cref{app:representations_permuted}).\looseness-1

\subsection{Linear Mixed-Effects Models}\label{subsec:lmm}
We extend \cref{eqn:predreadingtime} to account for subject- and document-level variability by fitting linear mixed-effects models (LMMs) with random intercepts for participants and documents on the MECO data; for high-dimensional representations, we first reduce the representations to 25 principal components. We selected the number of principal components using the scree-plot elbow criterion \citep{Cattell01041966}.
The results (\Cref{tab:lmm_scalar}) are broadly consistent with our main analyses, and significance over permuted reading times is preserved. At the same threshold $(\significance = 0.001)$, however, fewer predictors significantly outperform the baseline. This may follow from LMMs yielding more conservative effect estimates: by explicitly modeling reader and document variability, they isolate the predictor's contribution net of these sources, which simpler models can conflate into the effect itself.\looseness-1

\section{Discussion}
We now turn to discussing our results and their main implications. Across predictors (representations, information value, and logit-lens surprisal), we find substantial differences in their predictive power when extracted from different model layers. 

\paragraph{Opposite Trends for Logit Lens.} \citet{kuribayashi2025large} report that logit-lens surprisal performs better at earlier layers for larger models and at later layers for smaller ones. We corroborate this for mGPT (1.3B) and GPT-2 (117M), where logit-lens surprisal peaks at later layers. However, representations show the opposite pattern---higher predictive power at earlier layers---indicating that psychometric predictive power depends on the choice of predictor, not just model size. This contrast should be interpreted with care: representations are evaluated via linear regression on raw vectors, whereas logit-lens predictors undergo additional transformations (layer normalization, projection into vocabulary space), so the difference may partly reflect accessibility to a linear decoder rather than representational content alone. The superiority of early-layer representations for early-pass measures is consistent with the view that initial reading stages rely on lexical and morpho-syntactic features preserved in earlier layers.\footnote{More broadly, this pattern suggests a functional alignment between the layer hierarchy of the transformer and the temporal stages of human reading---early layers correspond to early-pass measures, while later layers correspond to late-pass measures---reminiscent of the depth--time correspondence observed in neuroscience \citep{caucheteux2022brains}.}

\paragraph{Cross-Lingual Differences.} On the MECO dataset, the performance of the different settings on first fixation duration is generally consistent between languages, as well as with the Provo data. 
However, we do not observe the same consistency across languages for gaze duration or total reading time, despite the fact that the non-baseline predictors are all derived from mGPT.\footnote{\citet{shliazhko-etal-2024-mgpt} report worse downstream performance for mGPT on Greek and Turkish, while \citet{arnett-bergen-2025-language} find that differences in downstream performance of language models seem to be caused by data quantity disparities, rather than model architecture.} 
In fact, \citet{kuribayashi2025large} also observed mixed results in multilingual settings, possibly due to latent effects of English on the processing of the target language \citep{wendler-etal-2024-llamas}. We conducted experiments with a monolingual Turkish model to investigate this possibility (\Cref{app:monolingual}); yet, we saw no discernible difference to the mGPT results for Turkish.
In a qualitative example of a particular clause in different languages (\Cref{app:meco_qualitative}), we observed some intra-lingual patterns that seem to modulate reading times: Greek seems more verbose and more likely to begin a structural unit with lower information content, because it is a language that favors the use of articles (e.g., even ahead of proper names). Turkish word order places verbs at the end of clauses. However, it is unclear if there is a connection between these patterns and the differences in the performance of representations on reading times.\looseness-1

\paragraph{Variance of Predictors across Layers.}
Finally, the validation MSE of both information value and logit-lens surprisal varies less across layers than representation-based predictors. One plausible reason is that they compress each layer's state into a single scalar, whereas representations expose a high-dimensional feature whose usefulness can change more sharply with layer depth.\looseness-1

\paragraph{Future Work.}
While this work provides a controlled comparison between several language-model-derived representations across layers and reading time measures, it leaves possible extensions for analyzing internal representations using dimensionality reduction or feature selection techniques.  In particular, kernel principal component analysis could be used to map representations into a non-linear feature space prior to probing, allowing us to assess whether reading-time-relevant structure is present but not linearly accessible in the original representations. Comparing layer-wise performance before and after such transformations would help disentangle representational content from linear decodability and clarify whether some of the observed trends are due to differences in how easily the information can be recovered by the probe, rather than differences in the information itself.
Moreover, our experiments are limited to mGPT, GPT-2, and cosmosGPT, and thus could be extended to other monolingual and/or larger language models.
Finally, we believe that the diverging performance of representations and logit-lens surprisal across layers offers interesting avenues to test with combinations of different predictors.\looseness-1

\section{Conclusion}
In this work, we investigated the psychometric power of language models by revisiting the hypothesis that reading times are best predicted by the scalar measure of surprisal by testing whether a neural language model's internal representations serve as more accurate predictors of human processing effort.
While controlling for standard psycholinguistic factors, we compared the predictive power of language model representations, information value, and layer-wise surprisal.
Across two eye-tracking corpora and five typologically distinct languages, we identify differences across reading time modalities: early-layer representations of mGPT and GPT-2 are superior at predicting early-pass measures (first fixation and gaze duration), while scalar surprisal remains superior for late-pass measures (total reading time). These results suggest that some psychometric power of language models is encoded within their internal representations beyond what surprisal captures. Notably, language model representations show more variability across layers compared to information value and logit-lens surprisal.
Finally, we find that the most effective predictor varies across the languages and eye-tracking measures analyzed.\looseness=-1

\section*{Limitations}

Our study is subject to several limitations. 
First, our analysis is restricted to eye-tracking data. While these measures provide high-fidelity temporal markers of reading effort, future work is needed to determine if the observed patterns generalize to other modalities, such as Self-Paced Reading or neuroimaging (EEG/fMRI). 
Second, computational constraints limited our evaluation to models up to 1.3B parameters (mGPT). It remains unclear whether the observed functional divergence between representations and surprisal persists or evolves in larger models with higher dimensionality. While prior work \citep{oh-schuler-2023-why, shain2024evidence, kuribayashi-etal-2024-psychometric} found that surprisal from smaller models often fits reading times better than that of larger language models, this scaling behavior may not generalize to a neural language model's internal representations.
Next, we acknowledge that probing experiments can lead to false discoveries if random noise in representations is not properly accounted for \citep{meloux2025everything}. While comparing our results against randomly initialized mGPT, GPT-2, and cosmosGPT baselines was not computationally feasible, we mitigate this concern through our permutation testing, which controls for random associations between predictors and reading times. Furthermore, we observe consistent trends across various experimental setups that random noise would not reproduce. However, we note that future research using randomly initialized model representations could examine our findings. In addition, our evaluation relies on raw MSE, which is scale-dependent: although this allows meaningful comparisons among predictors within the same eye-tracking measure, the magnitude of MSE differences should not be compared directly across first fixation duration, gaze duration, and total reading time, since these measures lie on different numerical scales.
Finally, we use a language model's raw internal representations without experimenting with dimensionality reduction methods, except in the case of mixed-effects models (\Cref{subsec:lmm}). Exploring lower-rank subspaces (e.g., via principal component analysis) could further isolate the specific features within the internal states that are accountable for the alignment with human processing effort. Relatedly, for multi-token units, we aggregate hidden states by mean-pooling; alternative aggregations (e.g., max-pool, first- or last-token pooling, or concatenation) may yield different results, and a systematic comparison is left to future work.\looseness=-1

\section*{Ethics Statement}
We foresee no ethical problems with our work.

\section*{Acknowledgments}
We would like to thank Alex Warstadt for helpful discussions, and Taiga Someya and Andreas Opedal for pointing us to \citet{kuribayashi2025large}. We also thank the
anonymous reviewers for their useful comments, suggestions, and references to related work. Eleftheria Tsipidi was supported by the SNSF grant number 204667. Karolina Sta\'nczak was supported by the ETH AI Center postdoctoral fellowship. We disclose the use of generative AI tools for light editing and rephrasing; the original text was our own, and we carefully reviewed all suggested edits.

\bibliography{custom}

\begin{thebibliography}{48}
\providecommand{\natexlab}[1]{#1}

\bibitem[{Alain and Bengio(2017)}]{alainbengio2017}
Guillaume Alain and Yoshua Bengio. 2017.
\newblock \href {https://arxiv.org/abs/1610.01644} {Understanding intermediate
  layers using linear classifier probes}.
\newblock In \emph{International Conference on Learning Representations}.

\bibitem[{Arnett and Bergen(2025)}]{arnett-bergen-2025-language}
Catherine Arnett and Benjamin Bergen. 2025.
\newblock \href {https://aclanthology.org/2025.coling-main.441/} {Why do
  language models perform worse for morphologically complex languages?}
\newblock In \emph{Proceedings of the International Conference on Computational
  Linguistics}.

\bibitem[{Belrose et~al.(2025)Belrose, Ostrovsky, McKinney, Furman, Smith,
  Halawi, Biderman, and Steinhardt}]{belrose2025tuned}
Nora Belrose, Igor Ostrovsky, Lev McKinney, Zach Furman, Logan Smith, Danny
  Halawi, Stella Biderman, and Jacob Steinhardt. 2025.
\newblock \href {https://arxiv.org/abs/2303.08112} {Eliciting latent
  predictions from transformers with the tuned lens}.

\bibitem[{Brown et~al.(2020)Brown, Mann, Ryder, Subbiah, Kaplan, Dhariwal,
  Neelakantan, Shyam, Sastry, Askell, Agarwal, Herbert-Voss, Krueger, Henighan,
  Child, Ramesh, Ziegler, Wu, Winter, Hesse, Chen, Sigler, Litwin, Gray, Chess,
  Clark, Berner, McCandlish, Radford, Sutskever, and
  Amodei}]{brown-etal-2020-gpt3}
Tom Brown, Benjamin Mann, Nick Ryder, Melanie Subbiah, Jared~D Kaplan, Prafulla
  Dhariwal, Arvind Neelakantan, Pranav Shyam, Girish Sastry, Amanda Askell,
  Sandhini Agarwal, Ariel Herbert-Voss, Gretchen Krueger, Tom Henighan, Rewon
  Child, Aditya Ramesh, Daniel Ziegler, Jeffrey Wu, Clemens Winter, and 12
  others. 2020.
\newblock \href
  {https://proceedings.neurips.cc/paper_files/paper/2020/file/1457c0d6bfcb4967418bfb8ac142f64a-Paper.pdf}
  {Language models are few-shot learners}.
\newblock In \emph{Advances in Neural Information Processing Systems},
  volume~33.

\bibitem[{Cattell(1966)}]{Cattell01041966}
Raymond~B. Cattell. 1966.
\newblock \href {https://doi.org/10.1207/s15327906mbr0102\_10} {The scree test
  for the number of factors}.
\newblock \emph{Multivariate Behavioral Research}, 1(2):245--276.
\newblock PMID: 26828106.

\bibitem[{Caucheteux and King(2022)}]{caucheteux2022brains}
Charlotte Caucheteux and Jean-R{\'e}mi King. 2022.
\newblock \href {https://doi.org/10.1038/s42003-022-03036-1} {Brains and
  algorithms partially converge in natural language processing}.
\newblock \emph{Communications Biology}, 5(1).

\bibitem[{Clifton et~al.(2007)Clifton, Staub, and Rayner}]{CLIFTON2007341}
Charles Clifton, Adrian Staub, and Keith Rayner. 2007.
\newblock \href
  {https://www.sciencedirect.com/science/article/pii/B9780080449807500173} {Eye
  movements in reading words and sentences}.
\newblock In \emph{Eye Movements}. Elsevier.

\bibitem[{Cook and Wei(2019)}]{cook2019eye}
Anne~E. Cook and Wencl Wei. 2019.
\newblock \href {https://doi.org/10.3390/vision3030045} {What can eye movements
  tell us about higher level comprehension?}
\newblock \emph{Vision}, 3(3).

\bibitem[{Gastaldi et~al.(2025)Gastaldi, Terilla, Malagutti, DuSell, Vieira,
  and Cotterell}]{gastaldi2025the}
Juan~Luis Gastaldi, John Terilla, Luca Malagutti, Brian DuSell, Tim Vieira, and
  Ryan Cotterell. 2025.
\newblock \href {https://openreview.net/forum?id=B5iOSxM2I0} {The foundations
  of tokenization: Statistical and computational concerns}.
\newblock In \emph{The International Conference on Learning Representations}.

\bibitem[{Giulianelli et~al.(2024{\natexlab{a}})Giulianelli, Malagutti,
  Gastaldi, DuSell, Vieira, and Cotterell}]{giulianelli-etal-2024-proper}
Mario Giulianelli, Luca Malagutti, Juan~Luis Gastaldi, Brian DuSell, Tim
  Vieira, and Ryan Cotterell. 2024{\natexlab{a}}.
\newblock \href {https://aclanthology.org/2024.emnlp-main.1032/} {On the proper
  treatment of tokenization in psycholinguistics}.
\newblock In \emph{Proceedings of the 2024 Conference on Empirical Methods in
  Natural Language Processing}.

\bibitem[{Giulianelli et~al.(2024{\natexlab{b}})Giulianelli, Opedal, and
  Cotterell}]{giulianelli-etal-2024-generalized}
Mario Giulianelli, Andreas Opedal, and Ryan Cotterell. 2024{\natexlab{b}}.
\newblock \href {https://aclanthology.org/2024.findings-emnlp.682/}
  {Generalized measures of anticipation and responsivity in online language
  processing}.
\newblock In \emph{Findings of the Association for Computational Linguistics:
  EMNLP 2024}.

\bibitem[{Giulianelli et~al.(2026)Giulianelli, Wallbridge, Cotterell, and
  Fern{\'a}ndez}]{giulianelli-etal-2024-incremental}
Mario Giulianelli, Sarenne Wallbridge, Ryan Cotterell, and Raquel
  Fern{\'a}ndez. 2026.
\newblock \href
  {https://www.sciencedirect.com/science/article/pii/S0749596X25001081}
  {Incremental alternative sampling as a lens into the temporal and
  representational resolution of linguistic prediction}.
\newblock \emph{Journal of Memory and Language}, 148.

\bibitem[{Giulianelli et~al.(2023)Giulianelli, Wallbridge, and
  Fern{\'a}ndez}]{giulianelli-etal-2023-information}
Mario Giulianelli, Sarenne Wallbridge, and Raquel Fern{\'a}ndez. 2023.
\newblock \href {https://aclanthology.org/2023.emnlp-main.343/} {Information
  value: Measuring utterance predictability as distance from plausible
  alternatives}.
\newblock In \emph{Proceedings of the Conference on Empirical Methods in
  Natural Language Processing}.

\bibitem[{Goodkind and Bicknell(2018)}]{goodkind2018predictive}
Adam Goodkind and Klinton Bicknell. 2018.
\newblock \href {https://aclanthology.org/W18-0102} {Predictive power of word
  surprisal for reading times is a linear function of language model quality}.
\newblock In \emph{Proceedings of the Workshop on Cognitive Modeling and
  Computational Linguistics}.

\bibitem[{Hale(2001)}]{hale-2001-probabilistic}
John Hale. 2001.
\newblock \href {https://aclanthology.org/N01-1021/} {A probabilistic {Earley}
  parser as a psycholinguistic model}.
\newblock In \emph{Proceedings of the Meeting of the North {American} Chapter
  of the Association for Computational Linguistics}.

\bibitem[{Immer et~al.(2022)Immer, Torroba~Hennigen, Fortuin, and
  Cotterell}]{immer-etal-2022-probing}
Alexander Immer, Lucas Torroba~Hennigen, Vincent Fortuin, and Ryan Cotterell.
  2022.
\newblock \href {https://aclanthology.org/2022.acl-long.129/} {Probing as
  quantifying inductive bias}.
\newblock In \emph{Proceedings of the Annual Meeting of the Association for
  Computational Linguistics (Volume 1: Long Papers)}.

\bibitem[{Just and Carpenter(1980)}]{just1980theory}
Marcel~A. Just and Patricia~A. Carpenter. 1980.
\newblock \href {https://doi.org/10.1037/0033-295X.87.4.329} {A theory of
  reading: From eye fixations to comprehension}.
\newblock \emph{Psychological Review}, 87(4).

\bibitem[{Kesgin et~al.(2024)Kesgin, Yuce, Dogan, Uzun, Uz, Seyrek, Zeer, and
  Amasyali}]{kesgin2024introducing}
H.~Toprak Kesgin, M.~Kaan Yuce, Eren Dogan, M.~Egemen Uzun, Atahan Uz, H.~Emre
  Seyrek, Ahmed Zeer, and M.~Fatih Amasyali. 2024.
\newblock \href {https://doi.org/10.1109/INISTA62901.2024.10683863}
  {Introducing {cosmosGPT}: Monolingual training for {Turkish} language
  models}.
\newblock In \emph{International Conference on INnovations in Intelligent
  SysTems and Applications (INISTA)}.

\bibitem[{Kiegeland et~al.(2026)Kiegeland, Sn{\ae}bjarnarson, Vieira, and
  Cotterell}]{kiegeland-etal-2026-units}
Samuel Kiegeland, V{\'e}steinn Sn{\ae}bjarnarson, Tim Vieira, and Ryan
  Cotterell. 2026.
\newblock On the proper treatment of units in surprisal theory.
\newblock In \emph{Proceedings of the Annual Meeting of the Association for
  Computational Linguistics}.

\bibitem[{Kim et~al.(2025)Kim, Evans, and Schein}]{kim2025linear}
Junsol Kim, James Evans, and Aaron Schein. 2025.
\newblock \href {https://arxiv.org/abs/2503.02080} {Linear representations of
  political perspective emerge in large language models}.
\newblock In \emph{The International Conference on Learning Representations}.

\bibitem[{Kuribayashi et~al.(2024)Kuribayashi, Oseki, and
  Baldwin}]{kuribayashi-etal-2024-psychometric}
Tatsuki Kuribayashi, Yohei Oseki, and Timothy Baldwin. 2024.
\newblock \href {https://aclanthology.org/2024.findings-naacl.129/}
  {Psychometric predictive power of large language models}.
\newblock In \emph{Findings of the Association for Computational Linguistics:
  NAACL 2024}.

\bibitem[{Kuribayashi et~al.(2025)Kuribayashi, Oseki, Taieb, Inui, and
  Baldwin}]{kuribayashi2025large}
Tatsuki Kuribayashi, Yohei Oseki, Souhaib~Ben Taieb, Kentaro Inui, and Timothy
  Baldwin. 2025.
\newblock \href {https://aclanthology.org/2025.tacl-1.78/} {Large language
  models are human-like internally}.
\newblock \emph{Transactions of the Association for Computational Linguistics},
  13.

\bibitem[{Levy(2008)}]{levy2008expectation}
Roger Levy. 2008.
\newblock \href
  {https://www.sciencedirect.com/science/article/pii/S0010027707001436}
  {Expectation-based syntactic comprehension}.
\newblock \emph{Cognition}, 106(3).

\bibitem[{Luke and Christianson(2018)}]{provo}
Steven~G. Luke and Kiel Christianson. 2018.
\newblock \href {https://link.springer.com/article/10.3758/s13428-017-0908-4}
  {The {Provo} corpus: A large eye-tracking corpus with predictability norms}.
\newblock \emph{Behavior Research Methods}, 50.

\bibitem[{Meister et~al.(2022)Meister, Pimentel, Clark, Cotterell, and
  Levy}]{meister-etal-2022-analyzing}
Clara Meister, Tiago Pimentel, Thomas Clark, Ryan Cotterell, and Roger Levy.
  2022.
\newblock \href {https://aclanthology.org/2022.acl-short.3/} {Analyzing wrap-up
  effects through an information-theoretic lens}.
\newblock In \emph{Proceedings of the Annual Meeting of the Association for
  Computational Linguistics (Volume 2: Short Papers)}.

\bibitem[{M{\'e}loux et~al.(2025)M{\'e}loux, Maniu, Portet, and
  Peyrard}]{meloux2025everything}
Maxime M{\'e}loux, Silviu Maniu, Fran{\c{c}}ois Portet, and Maxime Peyrard.
  2025.
\newblock \href {https://openreview.net/forum?id=5IWJBStfU7} {Everything,
  everywhere, all at once: Is mechanistic interpretability identifiable?}
\newblock In \emph{The International Conference on Learning Representations}.

\bibitem[{nostalgebraist(2020)}]{nostalgebraist2020logitlens}
nostalgebraist. 2020.
\newblock \href
  {https://www.lesswrong.com/posts/AcKRB8wDpdaN6v6ru/interpreting-gpt-the-logit-lens}
  {Interpreting {GPT}: The logit lens}.

\bibitem[{Oh and Schuler(2023)}]{oh-schuler-2023-why}
Byung-Doh Oh and William Schuler. 2023.
\newblock \href {https://doi.org/10.1162/tacl_a_00548} {Why does surprisal from
  larger transformer-based language models provide a poorer fit to human
  reading times?}
\newblock \emph{Transactions of the Association for Computational Linguistics},
  11.

\bibitem[{Oh and Schuler(2024)}]{oh-schuler-2024-leading}
Byung-Doh Oh and William Schuler. 2024.
\newblock \href {https://aclanthology.org/2024.emnlp-main.202/} {Leading
  whitespaces of language models' subword vocabulary pose a confound for
  calculating word probabilities}.
\newblock In \emph{Proceedings of the Conference on Empirical Methods in
  Natural Language Processing}.

\bibitem[{Opedal et~al.(2024)Opedal, Chodroff, Cotterell, and
  Wilcox}]{opedal-etal-2024-role}
Andreas Opedal, Eleanor Chodroff, Ryan Cotterell, and Ethan Wilcox. 2024.
\newblock \href {https://aclanthology.org/2024.emnlp-main.179/} {On the role of
  context in reading time prediction}.
\newblock In \emph{Proceedings of the Conference on Empirical Methods in
  Natural Language Processing}.

\bibitem[{Pimentel and Meister(2024)}]{pimentel-meister-2024-compute}
Tiago Pimentel and Clara Meister. 2024.
\newblock \href {https://aclanthology.org/2024.emnlp-main.1020/} {How to
  compute the probability of a word}.
\newblock In \emph{Proceedings of the Conference on Empirical Methods in
  Natural Language Processing}.

\bibitem[{Radford et~al.(2019)Radford, Wu, Child, Luan, Amodei, and
  Sutskever}]{radford2019language}
Alec Radford, Jeffrey Wu, Rewon Child, David Luan, Dario Amodei, and Ilya
  Sutskever. 2019.
\newblock \href
  {https://cdn.openai.com/better-language-models/language_models_are_unsupervised_multitask_learners.pdf}
  {Language models are unsupervised multitask learners}.

\bibitem[{Rayner(1998)}]{rayner1998eye}
Keith Rayner. 1998.
\newblock \href {http://dx.doi.org/10.1037/0033-2909.124.3.372} {Eye movements
  in reading and information processing: 20 years of research}.
\newblock \emph{Psychological Bulletin}, 124(3).

\bibitem[{Rayner(2009)}]{Rayner01082009}
Keith Rayner. 2009.
\newblock \href {https://doi.org/10.1080/17470210902816461} {Eye movements and
  attention in reading, scene perception, and visual search}.
\newblock \emph{The Quarterly Journal of Experimental Psychology}, 62(8).

\bibitem[{Rayner and Fischer(1996)}]{rayner1996mindless}
Keith Rayner and Martin~H. Fischer. 1996.
\newblock \href {https://doi.org/10.3758/BF03213106} {Mindless reading
  revisited: Eye movements during reading and scanning are different}.
\newblock \emph{Perception \& Psychophysics}, 58(5).

\bibitem[{Rayner et~al.(2000)Rayner, Kambe, and Duffy}]{rayner2000wrapup}
Keith Rayner, Gretchen Kambe, and Susan~A. Duffy. 2000.
\newblock \href {https://doi.org/10.1080/713755934} {The effect of clause
  wrap-up on eye movements during reading}.
\newblock \emph{The Quarterly Journal of Experimental Psychology Section A},
  53(4).

\bibitem[{Re et~al.(2025)Re, Opedal, Manaiev, Giulianelli, and
  Cotterell}]{re-etal-2025-spatio}
Francesco~Ignazio Re, Andreas Opedal, Glib Manaiev, Mario Giulianelli, and Ryan
  Cotterell. 2025.
\newblock \href {https://aclanthology.org/2025.acl-long.1474/} {A
  spatio-temporal point process for fine-grained modeling of reading behavior}.
\newblock In \emph{Proceedings of the Annual Meeting of the Association for
  Computational Linguistics (Volume 1: Long Papers)}.

\bibitem[{Schrimpf et~al.(2021)Schrimpf, Blank, Tuckute, Kauf, Hosseini,
  Kanwisher, Tenenbaum, and Fedorenko}]{schrimpf2021neural}
Martin Schrimpf, Idan~Asher Blank, Greta Tuckute, Carina Kauf, Eghbal~A.
  Hosseini, Nancy Kanwisher, Joshua~B. Tenenbaum, and Evelina Fedorenko. 2021.
\newblock \href {https://www.pnas.org/doi/abs/10.1073/pnas.2105646118} {The
  neural architecture of language: Integrative modeling converges on predictive
  processing}.
\newblock \emph{Proceedings of the National Academy of Sciences}, 118(45).

\bibitem[{Shain et~al.(2024)Shain, Meister, Pimentel, Cotterell, and
  Levy}]{shain2024evidence}
Cory Shain, Clara Meister, Tiago Pimentel, Ryan Cotterell, and Roger Levy.
  2024.
\newblock \href {https://www.pnas.org/doi/abs/10.1073/pnas.2307876121}
  {Large-scale evidence for logarithmic effects of word predictability on
  reading time}.
\newblock \emph{Proceedings of the National Academy of Sciences}, 121(10).

\bibitem[{Shliazhko et~al.(2024)Shliazhko, Fenogenova, Tikhonova, Kozlova,
  Mikhailov, and Shavrina}]{shliazhko-etal-2024-mgpt}
Oleh Shliazhko, Alena Fenogenova, Maria Tikhonova, Anastasia Kozlova, Vladislav
  Mikhailov, and Tatiana Shavrina. 2024.
\newblock \href {https://aclanthology.org/2024.tacl-1.4/} {m{GPT}: Few-shot
  learners go multilingual}.
\newblock \emph{Transactions of the Association for Computational Linguistics},
  12.

\bibitem[{Siegelman et~al.(2022)Siegelman, Schroeder, Acart{\"u}rk, Ahn,
  Alexeeva, Amenta, Bertram, Bonandrini, Brysbaert, and
  Chernova}]{siegelman2022expanding}
Noam Siegelman, Sascha Schroeder, Cengiz Acart{\"u}rk, Hee-Don Ahn, Svetlana
  Alexeeva, Simona Amenta, Raymond Bertram, Rolando Bonandrini, Marc Brysbaert,
  and Daria Chernova. 2022.
\newblock \href {https://link.springer.com/article/10.3758/s13428-021-01772-6}
  {Expanding horizons of cross-linguistic research on reading: {T}he
  {Multilingual Eye-movement Corpus} ({MECO})}.
\newblock \emph{Behavior Research Methods}, 54(6).

\bibitem[{Smith and Levy(2013)}]{SMITH2013302}
Nathaniel~J. Smith and Roger Levy. 2013.
\newblock \href
  {https://www.sciencedirect.com/science/article/pii/S0010027713000413} {The
  effect of word predictability on reading time is logarithmic}.
\newblock \emph{Cognition}, 128(3).

\bibitem[{Vaswani et~al.(2017)Vaswani, Shazeer, Parmar, Uszkoreit, Jones,
  Gomez, Kaiser, and Polosukhin}]{vaswani2017attention}
Ashish Vaswani, Noam Shazeer, Niki Parmar, Jakob Uszkoreit, Llion Jones,
  Aidan~N Gomez, {\L}ukasz Kaiser, and Illia Polosukhin. 2017.
\newblock \href
  {https://proceedings.neurips.cc/paper/2017/hash/3f5ee243547dee91fbd053c1c4a845aa-Abstract.html}
  {Attention is all you need}.
\newblock In \emph{Advances in Neural Information Processing Systems},
  volume~30.

\bibitem[{Vieira et~al.(2025)Vieira, Lebrun, Giulianelli, Gastaldi, Dusell,
  Terilla, O'Donnell, and Cotterell}]{pmlr-v267-vieira25a}
Tim Vieira, Benjamin Lebrun, Mario Giulianelli, Juan~Luis Gastaldi, Brian
  Dusell, John Terilla, Timothy~J. O'Donnell, and Ryan Cotterell. 2025.
\newblock \href {https://proceedings.mlr.press/v267/vieira25a.html} {From
  language models over tokens to language models over characters}.
\newblock In \emph{Proceedings of the International Conference on Machine
  Learning}, volume 267.

\bibitem[{Wendler et~al.(2024)Wendler, Veselovsky, Monea, and
  West}]{wendler-etal-2024-llamas}
Chris Wendler, Veniamin Veselovsky, Giovanni Monea, and Robert West. 2024.
\newblock \href {https://aclanthology.org/2024.acl-long.820/} {Do {Llamas} work
  in {English}? on the latent language of multilingual transformers}.
\newblock In \emph{Proceedings of the Annual Meeting of the Association for
  Computational Linguistics (Volume 1: Long Papers)}.

\bibitem[{White et~al.(2021)White, Pimentel, Saphra, and
  Cotterell}]{white-etal-2021-non}
Jennifer~C. White, Tiago Pimentel, Naomi Saphra, and Ryan Cotterell. 2021.
\newblock \href {https://aclanthology.org/2021.naacl-main.12/} {A non-linear
  structural probe}.
\newblock In \emph{Proceedings of the Conference of the North {American}
  Chapter of the Association for Computational Linguistics: Human Language
  Technologies}.

\bibitem[{Wilcox et~al.(2023)Wilcox, Pimentel, Meister, Cotterell, and
  Levy}]{wilcox-2023-testing}
Ethan~G. Wilcox, Tiago Pimentel, Clara Meister, Ryan Cotterell, and Roger~P.
  Levy. 2023.
\newblock \href {https://aclanthology.org/2023.tacl-1.82/} {Testing the
  predictions of surprisal theory in 11 languages}.
\newblock \emph{Transactions of the Association for Computational Linguistics},
  11.

\bibitem[{Wilcox et~al.(2020)Wilcox, Gauthier, Hu, Qian, and
  Levy}]{wilcox-etal:2020-on-the-predictive-power}
Ethan~Gotlieb Wilcox, Jon Gauthier, Jennifer Hu, Peng Qian, and Roger Levy.
  2020.
\newblock \href {https://arxiv.org/abs/2006.01912} {On the predictive power of
  neural language models for human real-time comprehension behavior}.
\newblock In \emph{Proceedings of the Cognitive Science Society}.

\end{thebibliography}

\clearpage

\appendix
\onecolumn

\section{Reproducibility}
\paragraph{Data Sources.} The license of both datasets is CC-By Attribution 4.0 International, which we adhere to. We obtained the Provo Corpus on the Open Science Framework.\footnote{\url{https://osf.io/sjefs}} We used the preprocessed version of the MECO data provided in \citet{opedal-etal-2024-role}.\footnote{\url{https://github.com/rycolab/context-reading-time/tree/main/merged_data_no_zero}}
\paragraph{Compute.} To estimate our predictors (surprisal, representations, information value, and logit lens), we used an RTX 2080 Ti GPU with 11GB VRAM for circa 30 hours. For our predictive modeling experiments, we used the same GPU, but reserving between 1--8GB of VRAM depending on the feature setting (with mGPT representations requiring the most compute), for approximately two months of compute time total.
\paragraph{Predictive Modeling.} We implemented our linear modeling using the \texttt{statsmodels}\footnote{\url{https://www.statsmodels.org/stable/index.html}} package. We attach the tuning hyperparameters in the supplementary material.

\newpage 

\section{Results for mGPT---Combined Settings}\label{app:combined}
\begin{figure*}[h!]
    \centering
    \includegraphics[width=\linewidth]{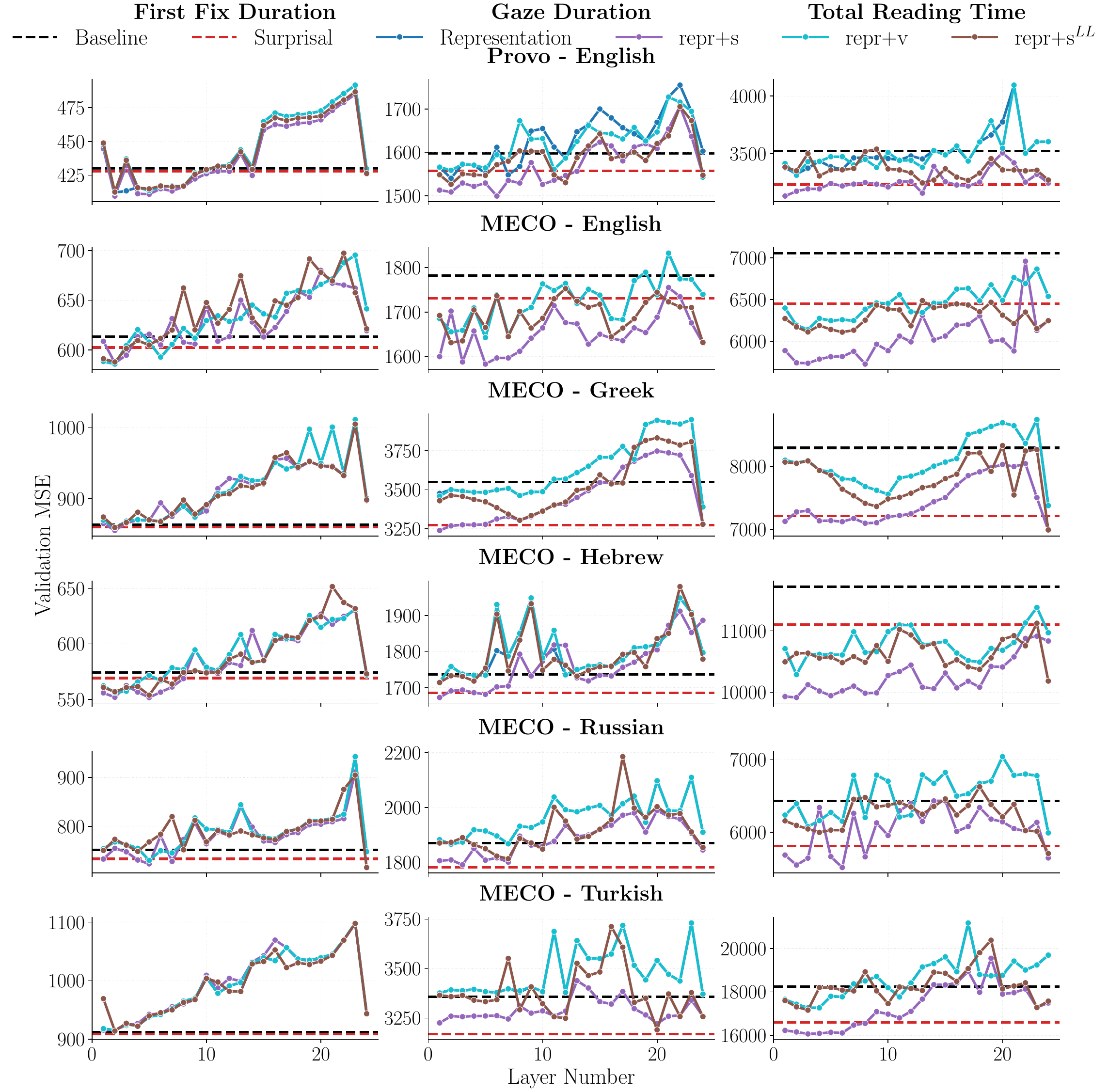}
    \caption{MSE for baseline, surprisal, and combined settings (representations with surprisal, information value, and logit-lens surprisal) on the Provo and MECO data across the 24 layers of mGPT and eye-tracking measures.}
    \label{fig:mse_mgpt_combo}
\end{figure*}

\clearpage
\begin{table*}[ht!]
\centering
\small
\setlength{\tabcolsep}{3pt}
\begin{tabular}{lrrrrrr}
\toprule
Measure & Surprisal & Best $\emb\;(\layer)$ & Best $\EmbS\;(\layer)$ & Best $\EmbV\;(\layer)$ & Best $\EmbLL\;(\layer)$ \\
\midrule
\multicolumn{6}{c}{\textbf{Provo---English}} \\
\midrule
FFD & -2.28$_{4.55}$$^{*}$ & -17.92$_{8.76}$$^{*\bullet}$ (2) & \textbf{-20.44$_{9.98}$$^{*\bullet}$ (2)} & -17.92$_{8.76}$$^{*\bullet}$ (2) & -17.46$_{8.50}$$^{*\bullet}$ (2) \\
GD & -39.96$_{34.34}$$^{*}$ & -57.21$_{31.77}$$^{*}$ (2) & \textbf{-98.01$_{85.12}$$^{*\bullet\ddagger}$ (6)} & -54.60$_{76.83}$$^{*\ddagger}$ (24) & -70.98$_{52.66}$$^{*}$ (2) \\
TRT & -290.61$_{134.19}$$^{*\bullet}$ & -198.09$_{196.67}$$^{*\bullet}$ (2) & \textbf{-389.00$_{167.39}$$^{*\bullet\ddagger}$ (1)} & -208.87$_{190.68}$$^{*\bullet}$ (2) & -275.41$_{214.61}$$^{*\bullet\ddagger}$ (13) \\
\midrule
\multicolumn{6}{c}{\textbf{MECO---English}} \\
\midrule
FFD & -11.09$_{18.93}$$^{*}$ & \textbf{-27.74$_{29.25}$$^{*}$ (2)} & -26.60$_{36.19}$$^{*}$ (2) & \textbf{-27.74$_{29.25}$$^{*}$ (2)} & -25.65$_{26.84}$$^{*\bullet}$ (2) \\
GD & -50.98$_{120.48}$$^{*}$ & -139.92$_{169.75}$$^{*\bullet}$ (5) & \textbf{-199.65$_{170.26}$$^{*\bullet}$ (5)} & -139.92$_{169.75}$$^{*\bullet}$ (5) & -151.29$_{142.44}$$^{*\bullet}$ (2) \\
TRT & -605.14$_{433.80}$$^{*\bullet}$ & -914.85$_{746.53}$$^{*\bullet}$ (3) & \textbf{-1327.41$_{976.18}$$^{*\bullet\ddagger}$ (8)} & -914.85$_{746.53}$$^{*\bullet}$ (3) & -944.99$_{857.93}$$^{*\bullet}$ (6) \\
\midrule
\multicolumn{6}{c}{\textbf{MECO---Greek}} \\
\midrule
FFD & -3.75$_{9.13}$$^{*}$ & -4.72$_{22.70}$$^{*}$ (2) & \textbf{-8.23$_{25.06}$$^{*}$ (2)} & -4.72$_{22.70}$$^{*}$ (2) & -4.09$_{26.86}$$^{*}$ (2) \\
GD & -275.80$_{212.98}$$^{*\bullet}$ & -158.94$_{306.51}$$^{*}$ (24) & \textbf{-309.45$_{269.98}$$^{*}$ (1)} & -158.95$_{306.52}$$^{*}$ (24) & -270.07$_{314.15}$$^{*}$ (24) \\
TRT & -1079.68$_{944.67}$$^{*\bullet}$ & -914.45$_{1072.60}$$^{*\bullet}$ (24) & \textbf{-1308.51$_{1378.06}$$^{*\bullet}$ (24)} & -914.60$_{1072.56}$$^{*\bullet}$ (24) & -1299.87$_{1316.34}$$^{*\bullet\ddagger}$ (24) \\
\midrule
\multicolumn{6}{c}{\textbf{MECO---Hebrew}} \\
\midrule
FFD & -5.04$_{5.08}$$^{*}$ & -18.31$_{24.28}$$^{*}$ (2) & \textbf{-22.56$_{22.10}$$^{*\ddagger}$ (5)} & -18.31$_{24.28}$$^{*}$ (2) & -20.45$_{27.61}$$^{*\ddagger}$ (5) \\
GD & -50.97$_{58.09}$$^{*}$ & -20.96$_{114.62}$$^{*}$ (1) & \textbf{-64.13$_{94.59}$$^{*}$ (1)} & -21.57$_{117.99}$$^{*}$ (1) & -22.71$_{116.94}$$^{*}$ (1) \\
TRT & -621.84$_{786.76}$$^{*\bullet}$ & -1431.52$_{2384.96}$$^{*\bullet}$ (2) & \textbf{-1804.89$_{1972.43}$$^{*\bullet}$ (2)} & -1431.52$_{2384.96}$$^{*\bullet}$ (2) & -1537.43$_{1621.02}$$^{*\bullet\ddagger}$ (24) \\
\midrule
\multicolumn{6}{c}{\textbf{MECO---Russian}} \\
\midrule
FFD & -18.74$_{22.90}$$^{\bullet}$ & -21.86$_{28.87}$$^{*}$ (5) & -36.11$_{49.96}$$^{*\ddagger}$ (24) & -21.86$_{28.87}$$^{*}$ (5) & \textbf{-36.37$_{49.17}$$^{*\ddagger}$ (24)} \\
GD & \textbf{-88.70$_{110.92}$$^{*\bullet}$} & -3.59$_{101.50}$$^{*}$ (2) & -78.96$_{175.41}$$^{*\ddagger}$ (3) & -3.59$_{101.50}$$^{*}$ (2) & -56.70$_{144.80}$$^{*}$ (7) \\
TRT & -619.68$_{741.79}$$^{*\bullet}$ & -438.41$_{605.05}$$^{*\bullet}$ (24) & \textbf{-914.07$_{1179.86}$$^{*\bullet\ddagger}$ (6)} & -438.61$_{605.01}$$^{*\bullet}$ (24) & -721.94$_{1213.85}$$^{*\bullet}$ (24) \\
\midrule
\multicolumn{6}{c}{\textbf{MECO---Turkish}} \\
\midrule
FFD & \textbf{-2.63$_{6.62}$} & 3.67$_{29.35}$$^{*}$ (2) & 4.01$_{31.17}$$^{*}$ (1) & 3.67$_{29.35}$$^{*}$ (2) & 2.35$_{32.29}$$^{*}$ (2) \\
GD & \textbf{-186.91$_{119.24}$$^{*\bullet}$} & 14.06$_{167.93}$$^{*}$ (24) & -134.56$_{163.52}$$^{*\ddagger}$ (20) & 13.96$_{167.93}$$^{*}$ (24) & -166.31$_{182.10}$$^{*\ddagger}$ (20) \\
TRT & -1642.88$_{1410.20}$$^{*\bullet}$ & -973.47$_{1871.74}$$^{*\bullet}$ (3) & \textbf{-2176.55$_{2190.16}$$^{*\bullet\ddagger}$ (3)} & -973.78$_{1871.72}$$^{*\bullet\ddagger}$ (3) & -1079.82$_{1906.71}$$^{*}$ (3) \\
\bottomrule
\end{tabular}
    \caption{$\dmse$ of ten-fold cross-validation for models trained on baseline features and mGPT-derived surprisal and representations, as well as combined settings: representations + surprisal ($\EmbS$), representations + information value ($\EmbV$), and representations + logit-lens surprisal ($\EmbLL$). Experiments were conducted on the Provo and MECO data across the 24 layers of mGPT and eye-tracking measures. For each measure, we report the lowest MSE over layers and the corresponding layer index $\layer$. Bold indicates the best condition per row. Asterisks (*) denote models that significantly outperform the respective models trained on permuted reading times, according to a one-sided paired $t$-test ($\significance = 0.001$). Similarly, bullets ($\bullet$) indicate significance over the baseline. In combined settings, double daggers ($\ddagger$) indicate significance over representation-trained models. None of the models in the combined settings were significant over their respective scalars, e.g. $\EmbS$ over surprisal.}
    \label{tab:mse_mgpt_combo}
\end{table*}

\clearpage
\section{Results for Monolingual Models}\label{app:monolingual}
\subsection{Individual Predictors}
\begin{figure*}[h!]
    \centering
    \includegraphics[width=\linewidth]{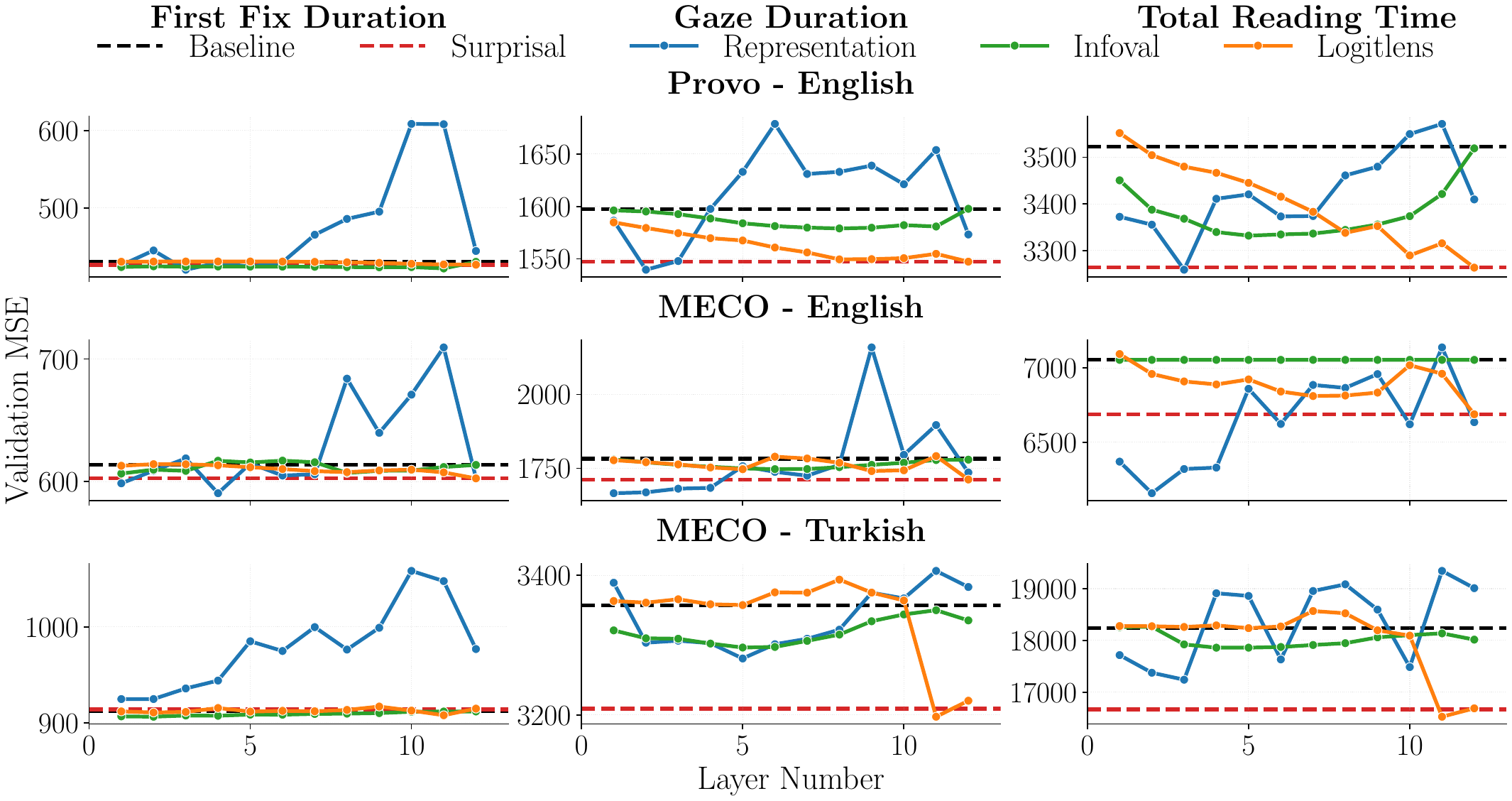}
    \caption{MSE for baseline, surprisal, representations, information value, and logit-lens surprisal on the Provo and English MECO with GPT-2 and Turkish MECO data with cosmosGPT, across the 12 layers of each language model and eye-tracking measures.}
    \label{fig:mse_gpt2_layers}
\end{figure*}

\begin{table*}[h!]
\centering
\begin{tabular}{lrrrrr}
\toprule
Measure & Surprisal & Best $\emb\;(\layer)$ & Best $\infoval\;(\layer)$ & Best $\logitlens\;(\layer)$ \\
\midrule
\multicolumn{5}{c}{\textbf{Provo---English}} \\
\midrule
FFD & -3.70$_{7.01}$$^{*}$ & \textbf{-10.05$_{13.84}$$^{*}$ (3)} & -8.39$_{11.53}$$^{*}$ (11) & -3.55$_{7.05}$$^{*}$ (12) \\
GD & -50.48$_{54.64}$$^{*}$ & \textbf{-57.92$_{90.88}$$^{*}$ (2)} & -18.51$_{28.90}$$^{*}$ (8) & -50.22$_{54.88}$$^{*}$ (12) \\
TRT & -258.98$_{205.34}$$^{*\bullet}$ & \textbf{-264.17$_{139.95}$$^{*\bullet}$ (3)} & -191.08$_{140.91}$$^{*\bullet}$ (5) & -259.69$_{199.76}$$^{*\bullet}$ (12) \\
\midrule
\multicolumn{5}{c}{\textbf{MECO---English}} \\
\midrule
FFD & -11.00$_{19.90}$$^{*}$ & \textbf{-23.12$_{19.55}$$^{*}$ (4)} & -7.06$_{11.95}$$^{*}$ (1) & -11.00$_{19.85}$$^{*}$ (12) \\
GD & -71.56$_{100.89}$$^{*}$ & \textbf{-116.72$_{136.39}$$^{*\bullet}$ (1)} & -35.20$_{41.36}$$^{*\bullet}$ (6) & -70.19$_{100.40}$$^{*}$ (12) \\
TRT & -366.37$_{445.23}$$^{*}$ & \textbf{-896.11$_{761.82}$$^{*\bullet}$ (2)} & 0.00$_{0.00}$$^{*}$ (1) & -365.45$_{439.37}$$^{*}$ (12) \\
\midrule
\multicolumn{5}{c}{\textbf{MECO---Turkish}} \\
\midrule
FFD & 2.20$_{13.84}$ & 12.91$_{24.99}$ (1) & \textbf{-5.51$_{7.51}$ (2)} & -3.97$_{6.28}$$^{*}$ (11) \\
GD & -148.33$_{78.18}$$^{*\bullet}$ & -76.27$_{196.86}$$^{*}$ (5) & -60.60$_{48.24}$$^{*}$ (5) & \textbf{-159.79$_{86.08}$$^{*\bullet}$ (11)} \\
TRT & -1575.56$_{1668.76}$$^{*\bullet}$ & -1000.99$_{1953.89}$$^{*\bullet}$ (3) & -381.99$_{319.72}$$^{*\bullet}$ (4) & \textbf{-1716.61$_{1877.91}$$^{*\bullet}$ (11)} \\
\bottomrule
\end{tabular}
    \caption{$\dmse$ (baseline $-$ target) of ten-fold cross-validation for models trained on surprisal, representations ($\emb$), information value ($\infoval$), and logit-lens surprisal ($\logitlens$) derived from GPT-2 for the Provo and English MECO data, and from cosmosGPT for Turkish MECO data, across the 12 embedding layers of each language model and eye-tracking measures. For each measure, we report the lowest MSE over layers and the corresponding layer index $\layer$. Bold indicates the best condition per row. Asterisks (*) denote models that significantly outperform the respective models trained on permuted reading times, according to a one-sided paired $t$-test ($\significance = 0.001$). Similarly, bullets ($\bullet$) indicate significance over the baseline.}
    \label{tab:mse_gpt2}
\end{table*}

\clearpage
\subsection{Combined Settings}
\begin{figure*}[h!]
    \centering
    \includegraphics[width=\linewidth]{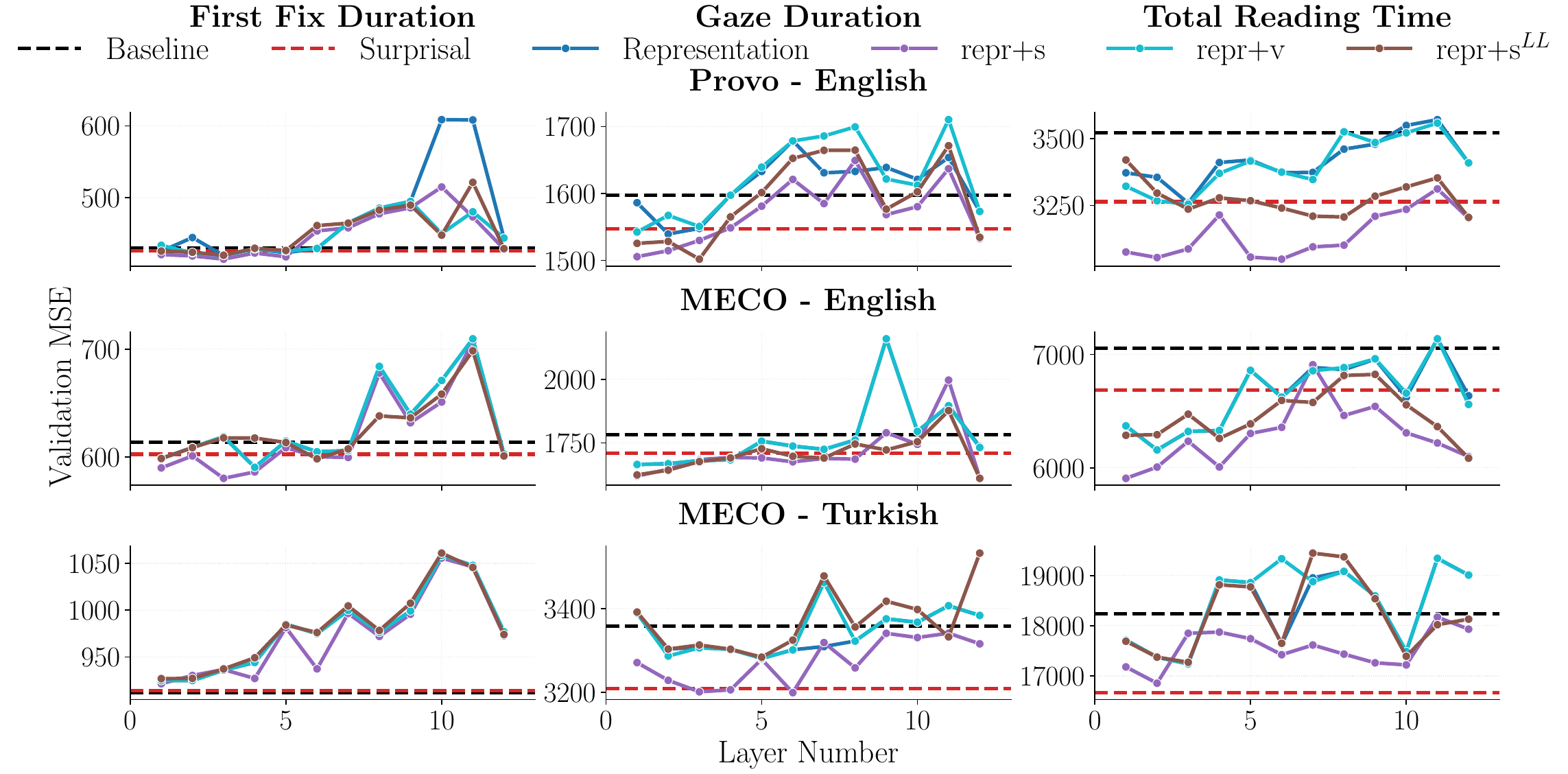}
    \caption{MSE for baseline, surprisal, and combined settings (representations with surprisal, information value, and logit-lens surprisal) on the Provo and English MECO data with GPT-2 and Turkish MECO data with cosmosGPT, across the 12 layers of each language model and eye-tracking measures.}
    \label{fig:mse_gpt2_combo}
\end{figure*}

\begin{table*}[h!]
\centering
\small
\setlength{\tabcolsep}{3pt}
\begin{tabular}{lrrrrrr}
\toprule
Measure & Surprisal & Best $\emb\;(\layer)$ & Best $\EmbS\;(\layer)$ & Best $\EmbV\;(\layer)$ & Best $\EmbLL\;(\layer)$ \\
\midrule
\multicolumn{6}{c}{\textbf{Provo---English}} \\
\midrule
FFD & -3.70$_{7.01}$$^{*}$ & -10.05$_{13.84}$$^{*}$ (3) & \textbf{-15.18$_{19.58}$$^{*}$ (3)} & -10.05$_{13.84}$$^{*}$ (3) & -9.77$_{14.20}$$^{*}$ (3) \\
GD & -50.48$_{54.64}$$^{*}$ & -57.92$_{90.88}$$^{*}$ (2) & -91.86$_{96.53}$$^{*\ddagger}$ (1) & -54.96$_{76.80}$$^{*}$ (1) & \textbf{-95.54$_{117.95}$$^{*}$ (3)} \\
TRT & -258.98$_{205.34}$$^{*\bullet}$ & -264.17$_{139.95}$$^{*\bullet}$ (3) & \textbf{-474.29$_{291.50}$$^{*\bullet\ddagger}$ (6)} & -268.57$_{135.91}$$^{*\bullet}$ (3) & -317.86$_{292.19}$$^{*\bullet\ddagger}$ (12) \\
\midrule
\multicolumn{6}{c}{\textbf{MECO---English}} \\
\midrule
FFD & -11.00$_{19.90}$$^{*}$ & -23.12$_{19.55}$$^{*}$ (4) & \textbf{-33.29$_{36.91}$$^{*\bullet\ddagger}$ (3)} & -23.12$_{19.55}$$^{*}$ (4) & -15.20$_{27.29}$$^{*}$ (6) \\
GD & -71.56$_{100.89}$$^{*}$ & -116.72$_{136.39}$$^{*\bullet}$ (1) & -168.44$_{176.21}$$^{*\bullet\ddagger}$ (12) & -116.72$_{136.39}$$^{*\bullet}$ (1) & \textbf{-171.10$_{176.75}$$^{*\bullet\ddagger}$ (12)} \\
TRT & -366.37$_{445.23}$$^{*}$ & -896.11$_{761.82}$$^{*\bullet}$ (2) & \textbf{-1146.38$_{1069.94}$$^{*\bullet\ddagger}$ (1)} & -896.11$_{761.82}$$^{*\bullet}$ (2) & -968.82$_{822.72}$$^{*\bullet\ddagger}$ (12) \\
\midrule
\multicolumn{6}{c}{\textbf{MECO---Turkish}} \\
\midrule
FFD & 2.20$_{13.84}$ & 12.91$_{24.99}$ (1) & 9.50$_{23.80}$ (1) & 12.91$_{24.99}$ (1) & 14.96$_{26.12}$ (1) \\
GD & -148.33$_{78.18}$$^{*\bullet}$ & -76.27$_{196.86}$$^{*}$ (5) & \textbf{-158.12$_{176.68}$$^{*}$ (6)} & -76.55$_{196.89}$$^{*\ddagger}$ (5) & -73.45$_{199.55}$$^{*}$ (5) \\
TRT & \textbf{-1575.56$_{1668.76}$$^{*\bullet}$} & -1000.99$_{1953.89}$$^{*\bullet}$ (3) & -1388.19$_{3077.40}$$^{*}$ (2) & -1003.79$_{1953.62}$$^{*\bullet\ddagger}$ (3) & -970.59$_{1945.13}$$^{*\bullet}$ (3) \\
\bottomrule
\end{tabular}
    \caption{As in \Cref{tab:mse_gpt2}, except we now consider the $\dmse$ of surprisal, representations, and combined settings: representations + surprisal ($\EmbS$), representations + information value ($\EmbV$), and representations + logit-lens surprisal ($\EmbLL$). In combined settings, double daggers ($\ddagger$) indicate statistical significance over models trained on representations. Note that all of these predictors performed worse than the baseline for first fixation duration on Turkish.}
    \label{tab:gpt2_combo}
\end{table*}

\clearpage
\section{Qualitative Example}\label{app:meco_qualitative}
\begin{figure*}[h!]
    \centering
    \includegraphics[width=\linewidth]{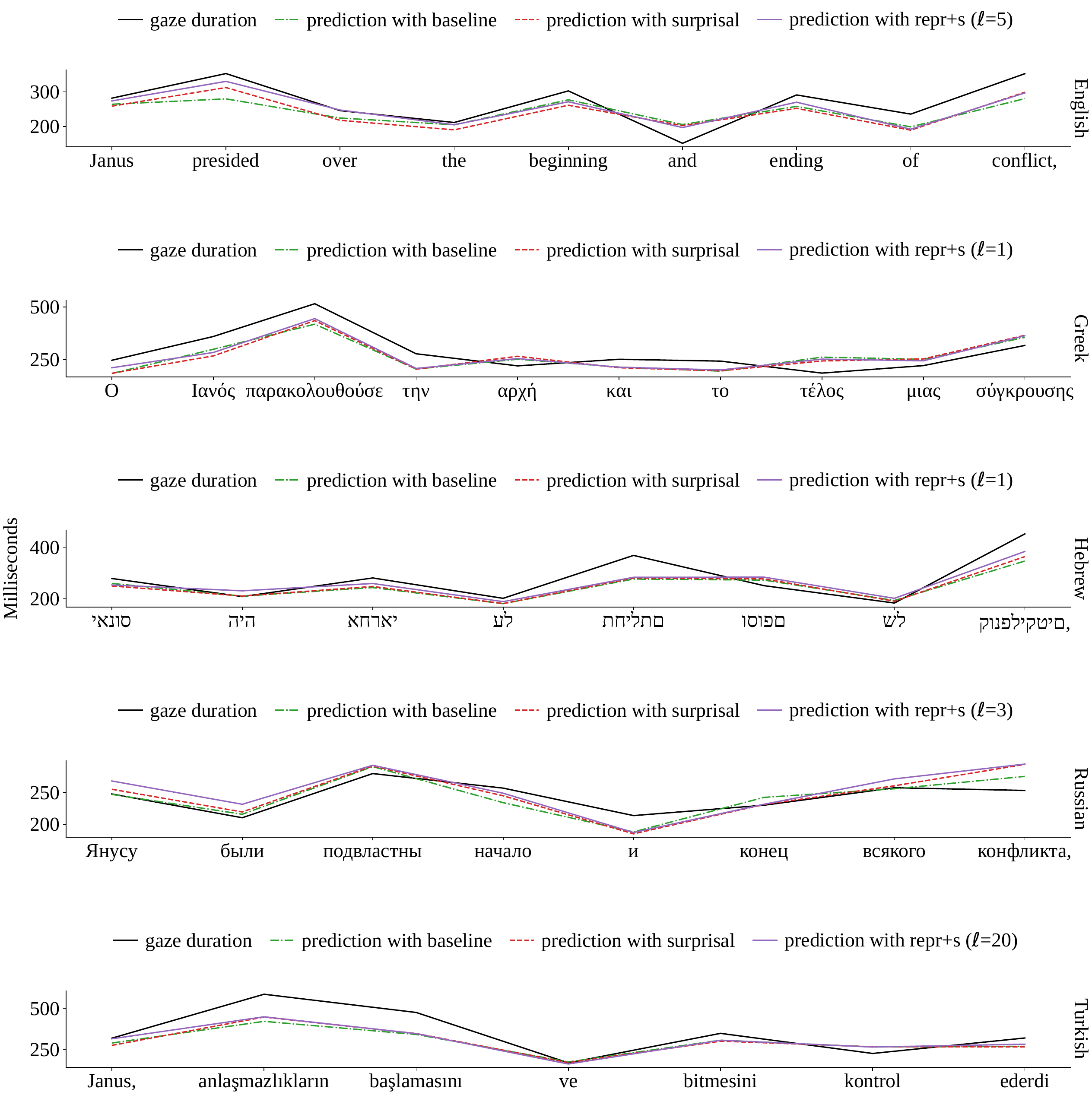}
    \caption{Gaze duration and its prediction by different mGPT-derived feature settings. We show the same excerpt from a document in the MECO dataset in different languages. The y-axis represents reading time measured in milliseconds. True gaze duration is represented by a black line. The purple line represents the prediction of a linear model trained on $\layer^{\text{th}}$ layer representations and standard surprisal. The chosen layer was the best for this feature setting and eye tracking measure per \Cref{tab:mse_mgpt_combo}. Note that Hebrew is in reverse order, as Hebrew is read and written right-to-left, and MECO data has words indexed by the order they are read.}
    \label{fig:meco_stacked}
\end{figure*}

\clearpage
\section{Permuted Results}\label{app:representations_permuted}
\subsection{mGPT---Permuted}
\subsubsection{Individual Predictors}
\begin{figure*}[h!]
    \centering
    \includegraphics[width=\linewidth]{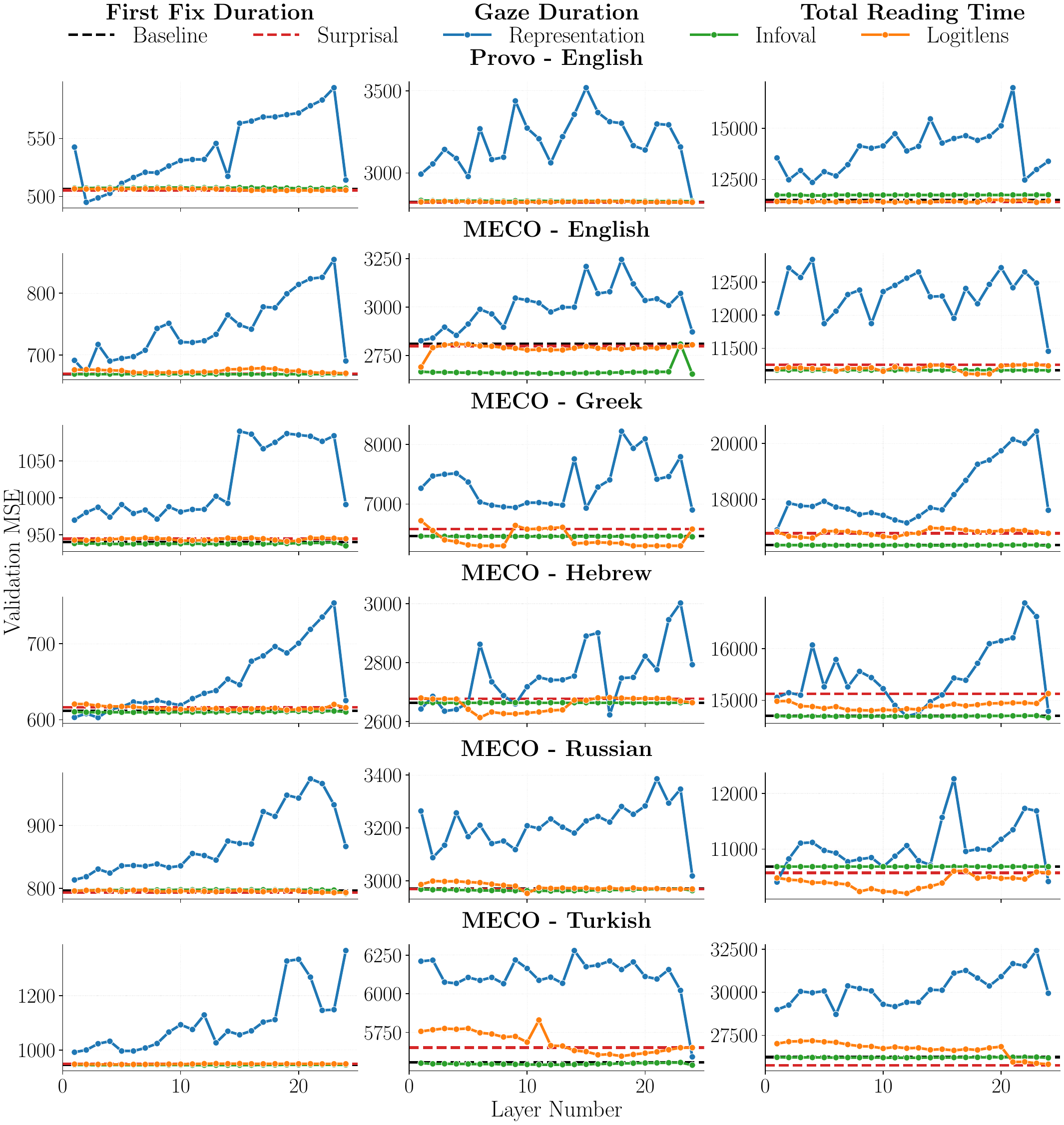}
    \caption{MSE for baseline, surprisal, representations, information value, and logit-lens surprisal on the Provo and MECO data across the 24 layers of mGPT and eye-tracking measures \textbf{with reading times randomly permuted during training}.}
    \label{fig:mse_mgpt_perm}
\end{figure*}

\clearpage
\begin{table*}[ht!]
\centering
\begin{tabular}{lrrrrr}
\toprule
Measure & Surprisal & Best $\emb\;(\layer)$ & Best $\infoval\;(\layer)$ & Best $\logitlens\;(\layer)$ \\
\midrule
\multicolumn{5}{c}{\textbf{Provo---English}} \\
\midrule
FFD & -0.79$_{2.96}$ & \textbf{-11.23$_{16.03}$$^{\bullet}$ (2)} & 0.80$_{4.66}$ (20) & -1.07$_{3.90}$ (18) \\
GD & \textbf{-2.64$_{10.05}$} & 2.87$_{158.05}$ (24) & 3.26$_{14.14}$ (22) & -2.61$_{10.06}$ (24) \\
TRT & -84.88$_{131.25}$$^{\bullet}$ & 869.44$_{286.65}$ (4) & 232.09$_{211.80}$ (4) & \textbf{-102.83$_{182.46}$$^{\bullet}$ (23)} \\
\midrule
\multicolumn{5}{c}{\textbf{MECO---English}} \\
\midrule
FFD & 0.47$_{5.30}$ & 1.30$_{31.42}$ (2) & 0.00$_{0.00}$ (1) & 1.27$_{3.26}$ (24) \\
GD & -9.90$_{31.13}$$^{\bullet}$ & 17.19$_{46.97}$ (1) & \textbf{-154.72$_{103.73}$$^{\bullet}$ (24)} & -118.38$_{102.49}$$^{\bullet}$ (1) \\
TRT & 81.39$_{185.30}$ & 284.92$_{441.58}$ (24) & 0.00$_{0.00}$ (1) & \textbf{-59.33$_{425.79}$ (18)} \\
\midrule
\multicolumn{5}{c}{\textbf{MECO---Greek}} \\
\midrule
FFD & 4.68$_{4.57}$ & 29.90$_{31.19}$ (1) & \textbf{-4.91$_{8.69}$ (24)} & 1.06$_{10.30}$ (19) \\
GD & 120.19$_{119.49}$ & 438.84$_{475.50}$ (24) & -11.17$_{20.89}$ (24) & \textbf{-162.05$_{111.87}$$^{\bullet}$ (23)} \\
TRT & 419.74$_{392.78}$ & 540.16$_{1144.00}$ (1) & \textbf{-30.47$_{28.26}$ (24)} & 247.26$_{667.37}$ (4) \\
\midrule
\multicolumn{5}{c}{\textbf{MECO---Hebrew}} \\
\midrule
FFD & 4.84$_{17.24}$ & \textbf{-8.71$_{27.72}$ (3)} & -1.84$_{1.72}$ (6) & 1.19$_{17.98}$ (19) \\
GD & 13.46$_{28.99}$ & -41.55$_{390.14}$ (17) & 0.00$_{0.00}$ (1) & \textbf{-50.71$_{111.73}$$^{\bullet}$ (6)} \\
TRT & 421.20$_{992.75}$ & -12.78$_{1579.43}$ (12) & \textbf{-38.06$_{38.82}$$^{\bullet}$ (24)} & 97.95$_{123.34}$ (9) \\
\midrule
\multicolumn{5}{c}{\textbf{MECO---Russian}} \\
\midrule
FFD & -3.28$_{7.87}$ & 16.77$_{23.60}$ (1) & \textbf{-4.63$_{6.13}$$^{\bullet}$ (24)} & -3.48$_{8.98}$ (23) \\
GD & -1.69$_{74.90}$ & 47.14$_{159.98}$ (24) & -9.45$_{10.76}$ (12) & \textbf{-18.68$_{84.95}$ (10)} \\
TRT & -111.87$_{260.18}$$^{\bullet}$ & -277.17$_{724.43}$ (1) & 0.00$_{0.00}$ (1) & \textbf{-480.24$_{638.05}$$^{\bullet}$ (12)} \\
\midrule
\multicolumn{5}{c}{\textbf{MECO---Turkish}} \\
\midrule
FFD & 2.90$_{4.52}$ & 45.91$_{30.86}$ (1) & \textbf{-1.71$_{3.66}$ (14)} & 0.92$_{4.37}$ (5) \\
GD & 95.13$_{99.69}$ & 35.26$_{407.39}$ (24) & \textbf{-18.35$_{22.87}$ (24)} & 39.64$_{48.78}$ (18) \\
TRT & \textbf{-469.13$_{886.55}$} & 2484.25$_{2163.78}$ (6) & -43.42$_{55.23}$ (24) & -418.17$_{887.85}$ (24) \\
\bottomrule
\end{tabular}
    \caption{$\dmse$ (baseline $-$ target) of ten-fold cross-validation for models trained on baseline features and mGPT-derived surprisal, representations ($\emb$), information value ($\infoval$), and logit-lens surprisal ($\logitlens$) \textbf{with reading times randomly permuted during training} on the Provo and MECO data across the 24 layers of mGPT. For each eye-tracking measure, we report the lowest MSE over layers and the corresponding layer index $\layer$. Bold indicates the best condition per row. Bullets ($\bullet$) denote models that significantly outperform the baseline, according to a one-sided paired $t$-test ($\significance = 0.001$).}
    \label{tab:perm_mse_mgpt}
\end{table*}

\clearpage
\subsubsection{Combined Settings}
\begin{figure*}[h!]
    \centering
    \includegraphics[width=\linewidth]{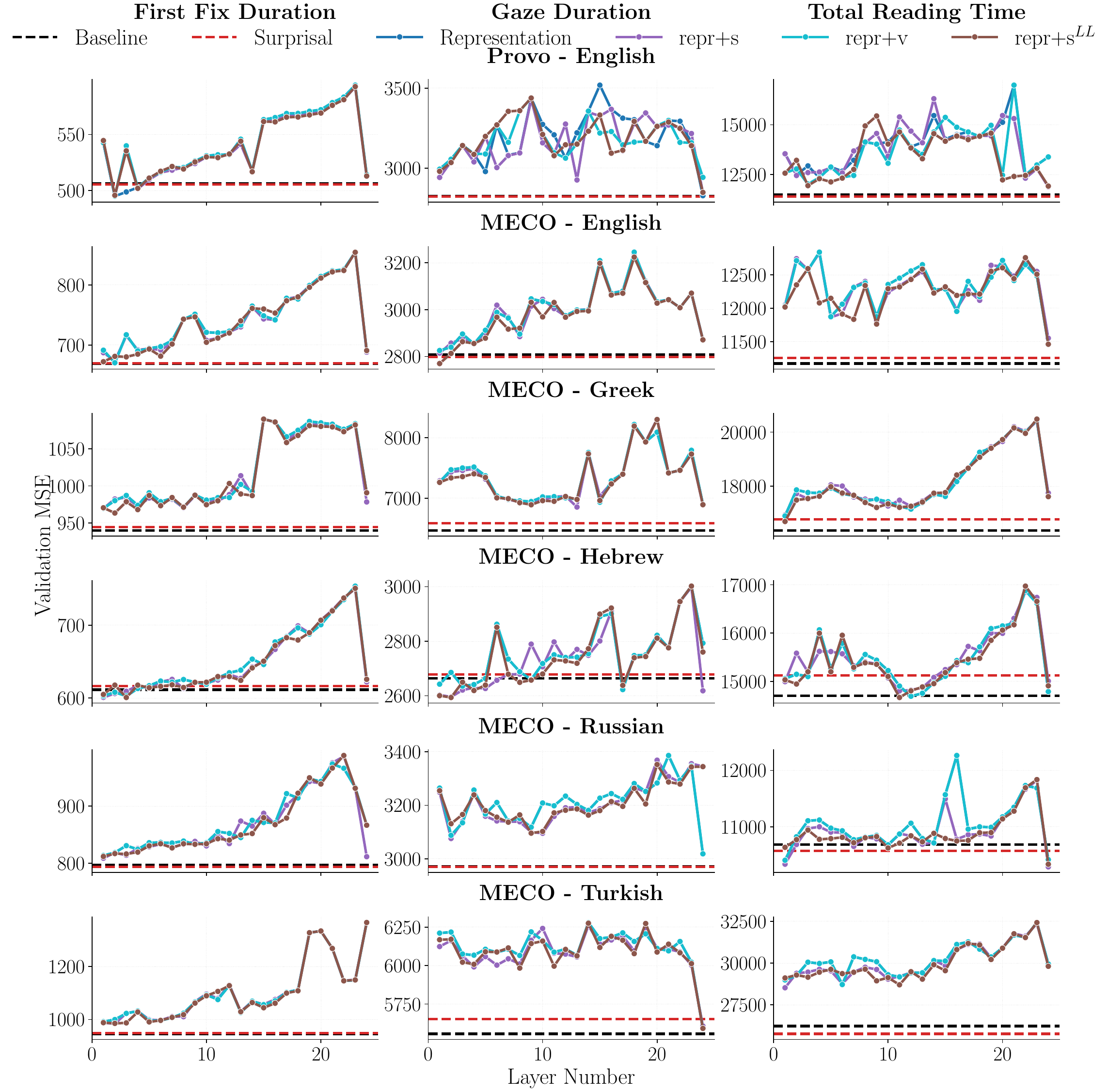}
    \caption{MSE for baseline, surprisal, and combined settings (representations with surprisal, information value, and logit-lens surprisal) on the Provo and MECO data across the 24 layers of mGPT and eye-tracking measures \textbf{with reading times randomly permuted during training}.}
    \label{fig:mse_mgpt_combo_perm}
\end{figure*}

\clearpage
\begin{table*}[ht!]
\centering
\small
\begin{tabular}{lrrrrrr}
\toprule
Measure & Surprisal & Best $\emb\;(\layer)$ & Best $\EmbS\;(\layer)$ & Best $\EmbV\;(\layer)$ & Best $\EmbLL\;(\layer)$ \\
\midrule
\multicolumn{6}{c}{\textbf{Provo---English}} \\
\midrule
FFD & -0.79$_{2.96}$ & -11.23$_{16.03}$$^{\bullet}$ (2) & \textbf{-11.53$_{16.29}$ (2)} & -11.23$_{16.03}$$^{\bullet}$ (2) & -10.44$_{15.46}$ (2) \\
GD & \textbf{-2.64$_{10.05}$} & 2.87$_{158.05}$ (24) & 22.04$_{87.40}$ (24) & 117.57$_{98.88}$ (24) & 23.41$_{86.33}$ (24) \\
TRT & \textbf{-84.88$_{131.25}$$^{\bullet}$} & 869.44$_{286.65}$ (4) & 399.78$_{508.09}$$^{\ddagger}$ (24) & 541.40$_{287.46}$$^{\ddagger}$ (3) & 438.03$_{572.60}$$^{\ddagger}$ (24) \\
\midrule
\multicolumn{6}{c}{\textbf{MECO---English}} \\
\midrule
FFD & 0.47$_{5.30}$ & 1.30$_{31.42}$ (2) & 4.55$_{26.14}$ (2) & 1.30$_{31.42}$ (2) & 3.09$_{17.74}$$^{\ddagger}$ (1) \\
GD & -9.90$_{31.13}$$^{\bullet}$ & 17.19$_{46.97}$ (1) & 5.79$_{50.99}$ (1) & 17.19$_{46.97}$ (1) & \textbf{-38.42$_{58.76}$ (1)} \\
TRT & 81.39$_{185.30}$ & 284.92$_{441.58}$ (24) & 379.85$_{481.71}$ (24) & 284.92$_{441.58}$ (24) & 289.60$_{447.32}$ (24) \\
\midrule
\multicolumn{6}{c}{\textbf{MECO---Greek}} \\
\midrule
FFD & 4.68$_{4.57}$ & 29.90$_{31.19}$ (1) & 29.76$_{29.49}$ (1) & 29.90$_{31.19}$ (1) & 23.61$_{29.14}$$^{\ddagger}$ (2) \\
GD & 120.19$_{119.49}$ & 438.84$_{475.50}$ (24) & 389.14$_{380.88}$ (13) & 438.84$_{475.50}$ (24) & 427.27$_{323.76}$ (9) \\
TRT & 419.74$_{392.78}$ & 540.16$_{1144.00}$ (1) & 576.88$_{1128.93}$ (1) & 540.16$_{1144.00}$ (1) & 333.44$_{648.44}$ (1) \\
\midrule
\multicolumn{6}{c}{\textbf{MECO---Hebrew}} \\
\midrule
FFD & 4.84$_{17.24}$ & -8.71$_{27.72}$ (3) & -10.58$_{29.66}$ (1) & -8.71$_{27.72}$ (3) & \textbf{-10.72$_{32.18}$ (3)} \\
GD & 13.46$_{28.99}$ & -41.55$_{390.14}$ (17) & -68.35$_{181.72}$$^{\ddagger}$ (2) & -41.55$_{390.14}$ (17) & \textbf{-70.53$_{166.65}$$^{\ddagger}$ (2)} \\
TRT & 421.20$_{992.75}$ & -12.78$_{1579.43}$ (12) & 99.43$_{1393.98}$ (11) & -12.78$_{1579.43}$ (12) & \textbf{-38.39$_{1493.58}$$^{\ddagger}$ (11)} \\
\midrule
\multicolumn{6}{c}{\textbf{MECO---Russian}} \\
\midrule
FFD & \textbf{-3.28$_{7.87}$} & 16.77$_{23.60}$ (1) & 11.91$_{25.19}$ (1) & 16.77$_{23.60}$ (1) & 15.07$_{22.85}$ (1) \\
GD & \textbf{-1.69$_{74.90}$} & 47.14$_{159.98}$ (24) & 104.39$_{103.45}$ (2) & 47.14$_{159.98}$ (24) & 124.32$_{196.86}$ (9) \\
TRT & -111.87$_{260.18}$$^{\bullet}$ & -277.17$_{724.43}$ (1) & \textbf{-395.25$_{1043.14}$$^{\bullet}$ (24)} & -277.17$_{724.43}$ (1) & -347.73$_{1064.00}$$^{\bullet}$ (24) \\
\midrule
\multicolumn{6}{c}{\textbf{MECO---Turkish}} \\
\midrule
FFD & 2.90$_{4.52}$ & 45.91$_{30.86}$ (1) & 38.61$_{32.40}$ (2) & 45.91$_{30.86}$ (1) & 39.56$_{40.33}$ (2) \\
GD & 95.13$_{99.69}$ & 35.26$_{407.39}$ (24) & 51.21$_{413.28}$ (24) & 35.26$_{407.39}$ (24) & 35.81$_{408.01}$ (24) \\
TRT & \textbf{-469.13$_{886.55}$} & 2484.25$_{2163.78}$ (6) & 2291.41$_{2137.59}$ (1) & 2484.25$_{2163.78}$ (6) & 2468.11$_{2508.89}$$^{\ddagger}$ (11) \\
\bottomrule
\end{tabular}
    \caption{$\dmse$ (baseline $-$ target) of ten-fold cross-validation for models trained on baseline features and mGPT-derived surprisal, as well as combined settings: representations + surprisal ($\EmbS$), representations + information value ($\EmbV$), and representations + logit-lens surprisal ($\EmbLL$), \textbf{with reading times randomly permuted during training} on the Provo and MECO data across the 24 layers of mGPT. For each eye-tracking measure, we report the lowest MSE over layers and the corresponding layer index $\layer$. Bold indicates the best condition per row. Bullets ($\bullet$) denote models that significantly outperform the baseline, according to a one-sided paired $t$-test ($\significance = 0.001$). In combined settings, double daggers ($\ddagger$) indicate statistical significance over models trained on representations.}
    \label{tab:perm_mse_mgpt_combo}
\end{table*}

\clearpage
\subsection{Monolingual Models---Permuted}
\subsubsection{Individual Predictors}
\begin{figure*}[h!]
    \centering
    \includegraphics[width=\linewidth]{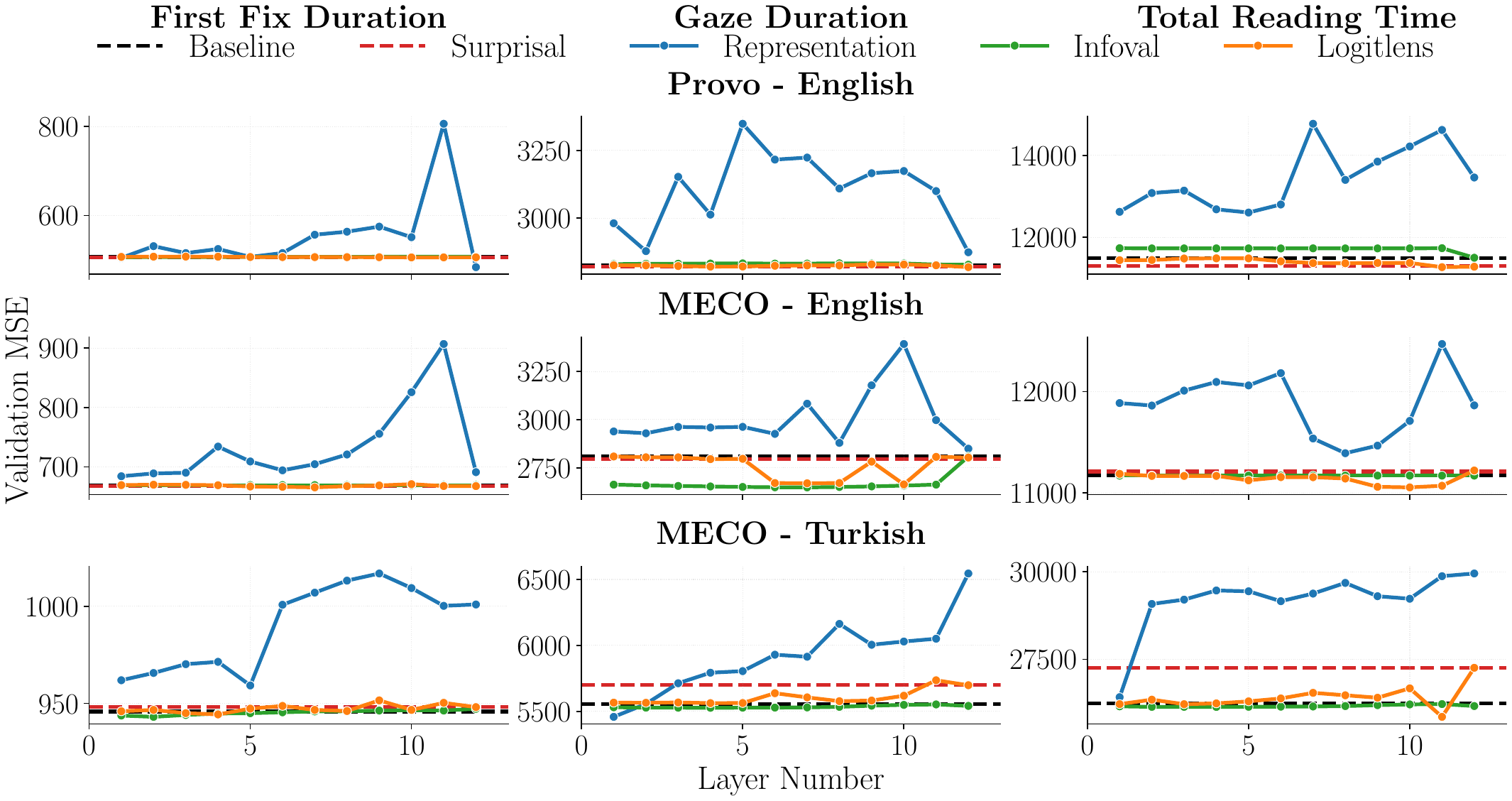}
    \caption{MSE for baseline, surprisal, representations, information value, and logit-lens surprisal on the Provo and English MECO with GPT-2 and Turkish MECO data with cosmosGPT, across the 12 layers of each language model and eye-tracking measures \textbf{with reading times randomly permuted during training}.}
    \label{fig:mse_gpt2_perm}
\end{figure*}

\begin{table*}[h!]
\centering
\begin{tabular}{lrrrrr}
\toprule
Measure & Surprisal & Best $\emb\;(\layer)$ & Best $\infoval\;(\layer)$ & Best $\logitlens\;(\layer)$ \\
\midrule
\multicolumn{5}{c}{\textbf{Provo---English}} \\
\midrule
FFD & -0.54$_{3.45}$ & \textbf{-22.56$_{18.65}$$^{\bullet}$ (12)} & -0.20$_{3.47}$ (1) & -0.61$_{3.40}$ (10) \\
GD & -5.04$_{11.92}$ & 49.20$_{91.78}$ (12) & 3.09$_{2.91}$ (12) & \textbf{-5.83$_{13.60}$ (12)} \\
TRT & -189.62$_{137.32}$$^{\bullet}$ & 1117.49$_{686.19}$ (5) & 14.75$_{160.53}$ (12) & \textbf{-214.38$_{170.94}$$^{\bullet}$ (11)} \\
\midrule
\multicolumn{5}{c}{\textbf{MECO---English}} \\
\midrule
FFD & -1.73$_{8.67}$ & 14.93$_{36.99}$ (1) & 0.00$_{0.00}$ (1) & \textbf{-3.79$_{7.83}$$^{\bullet}$ (7)} \\
GD & -11.95$_{30.49}$$^{\bullet}$ & 40.69$_{81.35}$ (12) & \textbf{-160.25$_{103.02}$$^{\bullet}$ (7)} & -144.79$_{100.08}$$^{\bullet}$ (10) \\
TRT & 40.57$_{151.86}$ & 221.43$_{666.71}$ (8) & 0.00$_{0.00}$ (1) & \textbf{-115.94$_{390.15}$$^{\bullet}$ (10)} \\
\midrule
\multicolumn{5}{c}{\textbf{MECO---Turkish}} \\
\midrule
FFD & 2.27$_{8.06}$ & 13.45$_{31.13}$ (5) & \textbf{-2.74$_{7.58}$ (2)} & -1.49$_{4.66}$ (4) \\
GD & 141.95$_{158.25}$ & \textbf{-98.55$_{123.02}$ (1)} & -29.89$_{28.11}$ (4) & 5.47$_{13.11}$ (4) \\
TRT & 1028.82$_{1222.60}$ & 184.14$_{976.70}$ (1) & -95.05$_{60.78}$$^{\bullet}$ (4) & \textbf{-378.28$_{822.30}$ (11)} \\
\bottomrule
\end{tabular}
    \caption{$\dmse$ (baseline $-$ target) of ten-fold cross-validation for models trained on baseline features and surprisal, representations ($\emb$), information value ($\infoval$), and logit-lens surprisal ($\logitlens$) derived from GPT-2 for the Provo and English MECO data, and from cosmosGPT for Turkish MECO data \textbf{with reading times randomly permuted during training}. For each measure, we report the lowest MSE over layers and the corresponding layer index $\layer$. Bold indicates the best condition per row. Bullets ($\bullet$) denote models that significantly outperform the baseline, according to a one-sided paired $t$-test ($\significance = 0.001$).}
    \label{tab:perm_mse_gpt2}
\end{table*}

\clearpage
\subsubsection{Combined Settings}
\begin{figure*}[h!]
    \centering
    \includegraphics[width=\linewidth]{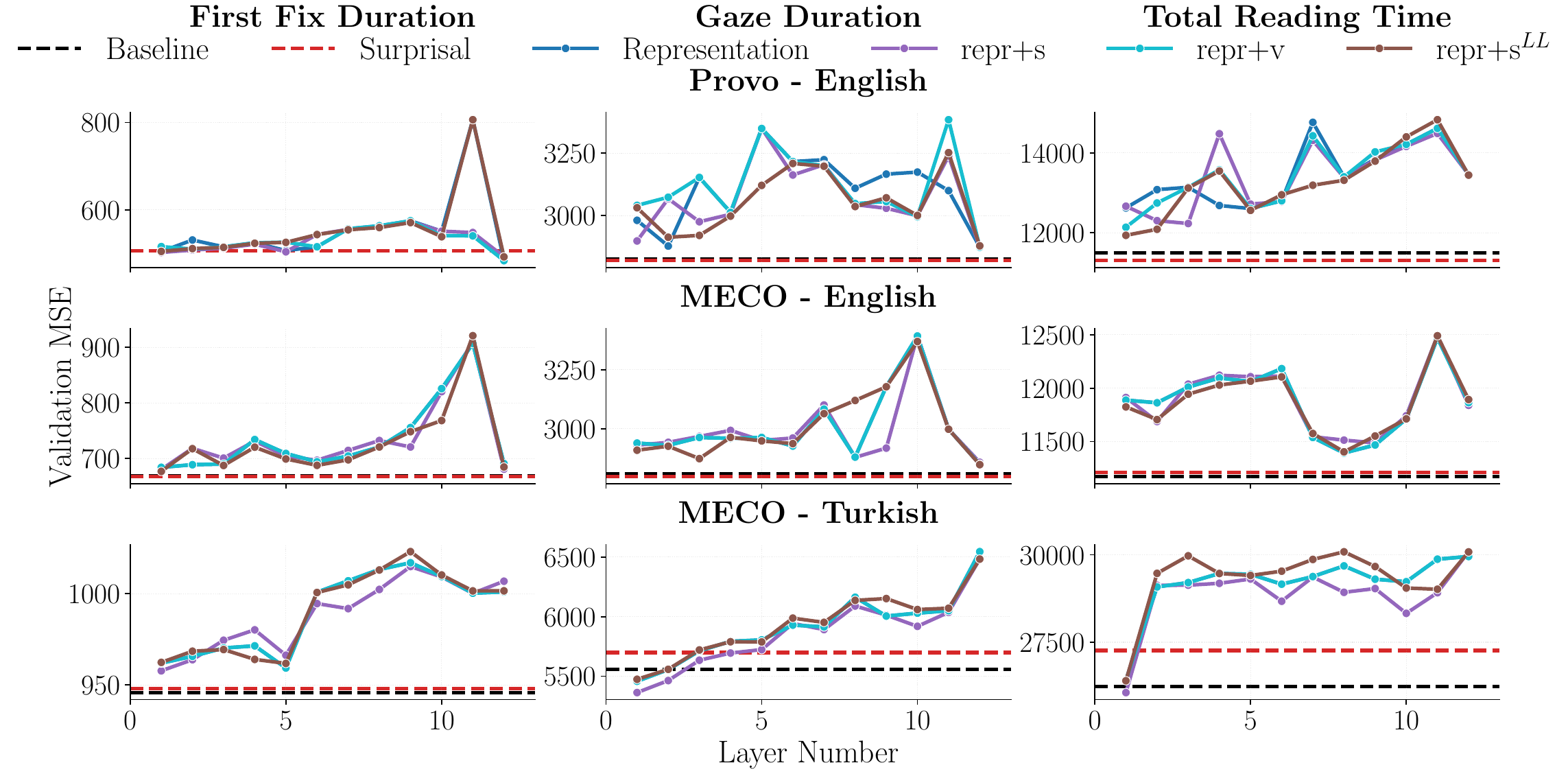}
    \caption{MSE for baseline, surprisal, and combined settings (representations with surprisal, information value, and logit-lens surprisal) on the Provo and English MECO data with GPT-2 and Turkish MECO data with cosmosGPT, across the 12 layers of each language model and eye-tracking measures \textbf{with reading times randomly permuted during training}.}
    \label{fig:mse_gpt2_combo_perm}
\end{figure*}

\begin{table*}[h!]
\centering
\small
\setlength{\tabcolsep}{5pt}
\begin{tabular}{lrrrrrr}
\toprule
Measure & Surprisal & Best $\emb\;(\layer)$ & Best $\EmbS\;(\layer)$ & Best $\EmbV\;(\layer)$ & Best $\EmbLL\;(\layer)$ \\
\midrule
\multicolumn{6}{c}{\textbf{Provo---English}} \\
\midrule
FFD & -0.54$_{3.45}$ & \textbf{-22.56$_{18.65}$$^{\bullet}$ (12)} & -13.52$_{16.36}$$^{\bullet}$ (12) & \textbf{-22.56$_{18.65}$$^{\bullet}$ (12)} & -13.91$_{16.25}$$^{\bullet}$ (12) \\
GD & \textbf{-5.04$_{11.92}$} & 49.20$_{91.78}$ (12) & 48.83$_{98.24}$ (12) & 49.20$_{91.78}$ (12) & 54.42$_{108.56}$ (12) \\
TRT & \textbf{-189.62$_{137.32}$$^{\bullet}$} & 1117.49$_{686.19}$ (5) & 742.37$_{531.21}$$^{\ddagger}$ (3) & 648.53$_{602.57}$$^{\ddagger}$ (1) & 446.26$_{503.46}$$^{\ddagger}$ (1) \\
\midrule
\multicolumn{6}{c}{\textbf{MECO---English}} \\
\midrule
FFD & \textbf{-1.73$_{8.67}$} & 14.93$_{36.99}$ (1) & 11.08$_{39.49}$$^{\ddagger}$ (1) & 14.93$_{36.99}$$^{\ddagger}$ (1) & 7.67$_{36.17}$ (1) \\
GD & \textbf{-11.95$_{30.49}$$^{\bullet}$} & 40.69$_{81.35}$ (12) & 48.68$_{83.21}$ (12) & 40.69$_{81.35}$ (12) & 38.91$_{76.28}$ (12) \\
TRT & 40.57$_{151.86}$ & 221.43$_{666.71}$ (8) & 316.13$_{817.56}$ (9) & 221.43$_{666.71}$ (8) & 231.97$_{700.13}$ (8) \\
\midrule
\multicolumn{6}{c}{\textbf{MECO---Turkish}} \\
\midrule
FFD & 2.27$_{8.06}$ & 13.45$_{31.13}$ (5) & 11.92$_{23.33}$ (1) & 13.45$_{31.13}$ (5) & 15.96$_{33.27}$ (5) \\
GD & 141.95$_{158.25}$ & -98.55$_{123.02}$ (1) & \textbf{-193.53$_{118.33}$$^{\bullet\ddagger}$ (1)} & -98.70$_{123.15}$ (1) & -82.34$_{150.56}$ (1) \\
TRT & 1028.82$_{1222.60}$ & 184.14$_{976.70}$ (1) & \textbf{-173.65$_{901.31}$$^{\ddagger}$ (1)} & 183.57$_{976.36}$ (1) & 164.47$_{1191.25}$ (1) \\
\bottomrule
\end{tabular}
    \caption{$\dmse$ (baseline $-$ target) of ten-fold cross-validation for models trained on baseline features and surprisal, as well as combined settings: representations + surprisal ($\EmbS$), representations + information value ($\EmbV$), and representations + logit-lens surprisal ($\EmbLL$) derived from GPT-2 for the Provo and English MECO data, and from cosmosGPT for Turkish MECO data \textbf{with reading times randomly permuted during training}. For each measure, we report the lowest MSE over layers and the corresponding layer index $\layer$. Bold indicates the best condition per row. Bullets ($\bullet$) denote models that significantly outperform the baseline, according to a one-sided paired $t$-test ($\significance = 0.001$). Similarly, for combined settings, double daggers ($\ddagger$) indicate significance over representation-trained models.}
    \label{tab:perm_mse_gpt2_combo}
\end{table*}

\clearpage
\section{Linear Mixed-Effects Models}\label{app:lmm}
\begin{table*}[h!]
\centering
\begin{tabular}{lrrrr}
\toprule
Measure & Surprisal & Best $\emb\;(\layer)$ & Best $\infoval\;(\layer)$ & Best $\logitlens\;(\layer)$ \\
\midrule
\multicolumn{5}{c}{\textbf{MECO---English}} \\
\midrule
FFD & -19.67$_{25.32}$$^{*}$ & \textbf{-37.84$_{38.14}$ (12)} & -18.90$_{20.13}$$^{*}$ (11) & -19.78$_{25.58}$$^{*}$ (24) \\
GD & -127.14$_{126.48}$$^{*}$ & \textbf{-150.98$_{147.34}$$^{*}$ (6)} & -65.36$_{48.64}$$^{*}$ (9) & -128.45$_{128.23}$$^{*}$ (24) \\
TRT & -695.30$_{613.28}$$^{*}$ & \textbf{-1029.41$_{1181.09}$$^{*}$ (11)} & -361.08$_{256.73}$$^{*\bullet}$ (8) & -700.19$_{622.56}$$^{*}$ (24) \\
\midrule
\multicolumn{5}{c}{\textbf{MECO---Greek}} \\
\midrule
FFD & -5.66$_{13.18}$$^{*}$ & \textbf{-13.01$_{28.70}$$^{*}$ (7)} & -8.59$_{7.77}$$^{*}$ (1) & -7.03$_{14.01}$$^{*}$ (23) \\
GD & -301.80$_{271.98}$$^{*}$ & \textbf{-306.80$_{196.41}$$^{*\bullet}$ (9)} & -36.21$_{39.22}$$^{*}$ (6) & -304.54$_{248.30}$$^{*}$ (23) \\
TRT & -1321.44$_{1322.95}$$^{*}$ & -1295.00$_{1436.95}$$^{*}$ (17) & -249.45$_{183.68}$$^{*}$ (6) & \textbf{-1332.82$_{1225.38}$$^{*}$ (23)} \\
\midrule
\multicolumn{5}{c}{\textbf{MECO---Hebrew}} \\
\midrule
FFD & -5.04$_{7.55}$$^{*}$ & \textbf{-16.51$_{24.18}$$^{*}$ (13)} & -5.07$_{5.21}$$^{*}$ (6) & -5.64$_{6.74}$$^{*}$ (19) \\
GD & -31.10$_{54.09}$$^{*}$ & \textbf{-55.32$_{79.07}$$^{*}$ (13)} & -14.76$_{14.48}$$^{*}$ (6) & -30.46$_{53.98}$$^{*}$ (24) \\
TRT & -361.63$_{708.45}$$^{*}$ & \textbf{-1395.11$_{1771.37}$$^{*}$ (13)} & -154.15$_{176.65}$$^{*}$ (2) & -361.52$_{714.79}$$^{*}$ (24) \\
\midrule
\multicolumn{5}{c}{\textbf{MECO---Russian}} \\
\midrule
FFD & -2.51$_{7.49}$ & 27.48$_{92.70}$ (4) & -2.60$_{6.62}$ (12) & \textbf{-3.19$_{7.67}$ (23)} \\
GD & -3.82$_{34.12}$$^{*}$ & \textbf{-3772.16$_{6778.05}$ (18)} & 3.25$_{11.56}$$^{*}$ (12) & -27.24$_{59.93}$$^{*}$ (8) \\
TRT & -126.33$_{327.15}$$^{*}$ & \textbf{-8452.38$_{17523.53}$ (18)} & -4.75$_{57.05}$$^{*}$ (7) & -190.15$_{333.11}$$^{*}$ (8) \\
\midrule
\multicolumn{5}{c}{\textbf{MECO---Turkish}} \\
\midrule
FFD & \textbf{-7.42$_{6.62}$$^{*}$} & -1.99$_{22.27}$$^{*}$ (9) & -5.54$_{7.92}$$^{*}$ (12) & -7.25$_{6.48}$$^{*}$ (24) \\
GD & \textbf{-238.29$_{137.52}$$^{*\bullet}$} & -149.02$_{133.32}$$^{*}$ (12) & -102.56$_{64.73}$$^{*\bullet}$ (9) & -236.52$_{139.32}$$^{*\bullet}$ (24) \\
TRT & \textbf{-1503.51$_{1502.65}$$^{*}$} & -642.16$_{1890.21}$$^{*}$ (18) & -583.09$_{474.45}$$^{*}$ (6) & -1494.48$_{1500.08}$$^{*}$ (24) \\
\bottomrule
\end{tabular}

    \caption{$\dmse$ of 10-fold cross-validation using linear mixed-effects models (LMMs) on per-participant MECO reading times with mGPT-derived surprisal, representations ($\emb$; PCA with $K{=}25$ components), logit-lens surprisal ($\logitlens$), and information value ($\infoval$). Unlike the main analyses (\cref{tab:mse_mgpt}), which use regularized regression on reading times averaged across participants, here we retain individual observations and fit LMMs with random intercepts for subjects and documents $\readingtime_{s,i} = \mathbf{x}_{i}^{\top}\boldsymbol{\beta} + b_s + u_i + \varepsilon_{s,i}$, where $b_s \sim \mathcal{N}(0,\sigma^2_{\text{subj}})$ and $u_i \sim \mathcal{N}(0,\sigma^2_{\text{item}})$. Models are fit by maximum likelihood with \texttt{lme4}; test-set predictions use fixed effects only. For each predictor type, we report the best-performing layer and its index $\layer$. Bold indicates the best condition per row. Asterisks (*) denote models that significantly outperform the respective models trained on permuted reading times, according to a one-sided paired $t$-test ($\significance = 0.001$). Bullets ($\bullet$) indicate significance over the baseline. Note that the representation results for Russian exhibit high variance across folds, likely due to overfitting of the 25 PCA components on the smaller Russian dataset.}
    \label{tab:lmm_scalar}
\end{table*}

\end{document}